\def\year{2022}\relax
\documentclass[letterpaper]{article} 
\usepackage[submission]{aaai23}  
\usepackage{times}  
\usepackage{helvet}  
\usepackage{courier}  
\usepackage[hyphens]{url}  
\usepackage{graphicx} 
\usepackage{natbib}  
\usepackage{caption} 
\usepackage{soul}
\usepackage{url}
\usepackage[utf8]{inputenc}
\usepackage{amsmath}
\usepackage{amsthm}
\usepackage{amsfonts}
\usepackage{booktabs}
\usepackage{algorithm}
\usepackage{algorithmic}
\usepackage{subfigure}
\usepackage{makecell}
\usepackage{algorithm}
\usepackage{algorithmic}

\usepackage{newfloat}
\usepackage{listings}
\DeclareCaptionStyle{ruled}{labelfont=normalfont,labelsep=colon,strut=off} 
\floatstyle{ruled}
\newfloat{listing}{tb}{lst}{}
\floatname{listing}{Listing}
\title{Reaching Consensus in Cooperative Multi-Agent \\ Reinforcement Learning with Goal Imagination}
\author{
    \equalcontrib{Liangzhou Wang},\textsuperscript{\rm 1}
    \equalcontrib{Kaiwen Zhu},\textsuperscript{\rm 1}
    \equalcontrib{Fengming Zhu},\textsuperscript{\rm 1}
    Xinghu Yao,\textsuperscript{\rm 1}
    Shujie Zhang,\textsuperscript{\rm 1}
    Deheng Ye,\textsuperscript{\rm 1}
    Haobo Fu,\textsuperscript{\rm 1}
    Qiang Fu,\textsuperscript{\rm 1}
    Wei Yang\textsuperscript{\rm 1}
}
\affiliations{
    \textsuperscript{\rm 1} Tencent AI Lab, Shenzhen, China \\
    \{allanlzwang, kevinltzhu, fridazhu, xinghuyao, shujiezhang, dericye, haobofu, leonfu, willyang\}@tencent.com
}

\usepackage{bibentry}

\begin{document}

\maketitle

\newcommand{\MN}{MAGI }

\begin{abstract}
Reaching consensus is key to multi-agent coordination.
To accomplish a cooperative task, agents need to coherently select optimal joint actions to maximize the team reward.
However, current cooperative multi-agent reinforcement learning (MARL) methods usually do not explicitly take consensus into consideration, which may cause miscoordination problem.
In this paper, we propose a model-based consensus mechanism to explicitly coordinate multiple agents.
The proposed \textit{Multi-agent Goal Imagination} (MAGI) framework guides agents to reach consensus with an \textit{imagined} common goal.
The common goal is an achievable state with high value, which is obtained by sampling from the distribution of future states.
We directly model this distribution with a  self-supervised generative model, thus alleviating the ``curse of dimensinality" problem induced by multi-agent multi-step policy rollout commonly used in model-based methods.
We show that such efficient consensus mechanism can guide all agents cooperatively reaching valuable future states. Results on Multi-agent Particle-Environments and Google Research Football environment demonstrate the superiority of MAGI in both sample efficiency and performance.

\end{abstract}

\section{Introduction}
Cooperative multi-agent reinforcement learning has demonstrated its promising ability to deal with many complicated real-world problems, such as coordination of autonomous vehicles \cite{cao2012overview}, control of multiple robots \cite{huttenrauch2017guided}, and management of network routing \cite{ye2015multi}.
In cooperative multi-agent systems, consensus among agents is important to successful coordination \cite{ren2005survey}. Particularly, agents need to choose action coherently to reach the optimal joint decision.
For example, in bidirectional traffic, keeping to one side (left or right) of the road when driving is an important consensus to traffic flow, and agents have to reach a consensus in terms of which side to conform to.

Early methods addressing the cooperative MARL problems consider other agents as part of environment and learn policy for each agent individually \cite{matignon2012independent,tampuu2017multiagent}.
However, as the environment dynamics becomes non-stationary, it is hard for agents to reach consensus and jointly make optimal decision \cite{gronauer2022multi}.

Recent approaches adopt the centralized training with decentralized execution (CTDE) framework to handle the non-stationary issue, such as MADDPG \cite{lowe2017multi} and a series of value decomposition methods \cite{sunehag2018value,rashid2018qmix,son2019qtran,peng2021facmac}. In the training phase, the centralized structure in these CTDE methods provides an implicit consensus mechanism for agents' policy learning. However, in the execution phase, agents still make decisions independently.
Hence, miscoordination may arise due to lack of consensus and sub-optimal solutions are found as a result \cite{gronauer2022multi}.


Hierarchical reinforcement learning methods introduce a high-level policy to guide low-level policy. This architecture can be used to construct a consensus mechanism to coordinate multi-agent policies.
Representative methods like feudal network \cite{vezhnevets2017feudal,ahilan2019feudal} generate a hidden state goal with high-level policy, while the low-level policies of agents are supposed to reach the goal state.
The high-level state goal can provide a consensus mechanism that helps multiple agents behave coherently towards the state with potential high value.
However, in multi-agent environments, how to generate the goal and use it to coordinate agents to reach consensus effectively and efficiently still remains an open challenge.



To address above challenges, we propose \emph{Multi-agent Goal Imagination (MAGI)}, a cooperative MARL framework that generates goals to achieve multi-agent consensus in an effective and efficient way.

MAGI constructs an explicit consensus mechanism for multi-agent coordination.
This consensus is an achievable and valuable future state, which is used as a common goal to guide multi-agent policies.
Agents make decisions conditioned not only on their local observation, but also on the common goal.
This goal encourages agents to cooperatively explore towards a common target state with potential high reward.
To generate such a common goal, MAGI adopts a model-based way.
Specifically, MAGI uses a self-supervised conditional variational auto-encoder (CVAE) \cite{sohn2015learning} to model the distribution of future states.
The agents' behavior will be implicitly embedded in the self-supervised CVAE, and therefore possible future states can be sampled while alleviating the ``curse of dimensionality'' problem introduced by policy rollout.
After that, a valuable future state from the distribution will be selected as the common goal, and we call this process goal imagination.

In summary, our main contributions are listed as follows:
\begin{itemize}
    \item
    We propose a novel consensus mechanism for cooperative MARL.
    This consensus mechanism provides an explicit goal to coordinate multiple agents effectively.
    \item
    We introduce an efficient model-based goal generation method, which avoids the multi-step rollout procedure commonly used in model-based methods.
    \item
    Empirical results on Multi-agent Particle-Environments and challenging Google Research Football environments demonstrate the superiority of MAGI in terms of both performance and sample efficiency.
\end{itemize}


\section{Related Work}
\textbf{Cooperative MARL:}
Based on the development of single agent reinforcement learning, early methods handle the multi-agent reinforcement problem by regarding each agent individually and training independent learners \cite{matignon2012independent,tampuu2017multiagent}.
However, treating other agents as part of environment will cause the non-stationary dynamics problem \cite{omidshafiei2017deep}, and many popular MARL methods adopt a centralized training with decentralized execution (CTDE) framework to attenuate this problem. Representative CTDE methods include value decomposition networks (VDN) \cite{sunehag2018value}, QMIX \cite{rashid2018qmix}, MADDPG \cite{lowe2017multi} and FACMAC \cite{peng2021facmac}.

The CTDE methods adopt a centralized structure during training, which can be regarded as an implicit consensus mechanism for multi-agent coordination. However, during execution agents make decisions independently without such centralized guidance, which may result in sub-optimal solutions due to mis-coordination \cite{gronauer2022multi}. 
Goal conditioned methods can be used to enhance multi-agent learning.
Cooperative multi-agent exploration (CMAE) \cite{liu2021cooperative} selects unexplored states as the goal to improve exploration.
Feudal network \cite{vezhnevets2017feudal,ahilan2019feudal} adopts a high-level policy generating the hidden state goal to guide the low-level policies.

Compared with these methods, MAGI builds an explicit consensus mechanism by generating a model-based common-goal to guide multi-agent policies, which coordinates multiple agents effectively and efficiently.

\noindent\textbf{Model-based RL:} 
Model-based reinforcement learning provides a more sample efficient framework by learning an environment model first and then planning optimal control upon it \cite{sutton1991dyna}.
Representative model-based methods such as MPC \cite{nagabandi2018neural} are based on the step-by-step prediction that predicts the next observation given the current state and action.
However, the multi-step compounding error limits the performance of model-based methods\cite{talvitie2014model}. Some recent approaches directly model the long-term dynamics to alleviate this problem \cite{mishra2017prediction,ke2019learning,krupnik2020multi}.
However, the growing joint action space still makes their action-dependent planning hard to scale with agent numbers in multi-agent environments.

Compared with above mentioned model-based methods, the goal generation procedure of MAGI is more efficient since the common goal generation and action selection is disentangled in MAGI.
The multi-agent behavior is implicitly embedded in the generative model, so MAGI can directly model long-horizon future state distribution.

\section{Preliminary}
A single agent reinforcement learning problem can be described as a Markov Decision Process (MDP) \cite{sutton2018reinforcement}.
A MDP is defined by the tuple ($\mathcal{S},\mathcal{A}, P, r, \rho_0,\gamma$),
where $\mathcal{S}$ is a finite set of states,
$\mathcal{A}$ is a finite set of actions,
$P: \mathcal{S}\times{A}\times{S}\mapsto[0,1]$ is the transition probability distribution,
$r: \mathcal{S}\times\mathcal{A} \mapsto \mathbb{R}$ is the reward function,
$\rho_0$ is the distribution of the initial state $s_0$, and $\gamma\in(0,1)$ is the discount factor.
The agent aims to maximize its total expected reward $R = \sum_{t=0}^{T}\gamma^t r^t$, where $T$ is the time horizon.

A multi-agent extension of MDP calls Markov games \cite{littman2001value}.
A Markov game for $N$ agents is defined by a set of global states $\mathcal{S}$, and a set of actions $\mathcal{A}_1, \cdots, \mathcal{A}_N$ for each agent.
To choose actions, each agent $i$ uses a stochastic policy $\pi_{\theta^i}:\mathcal{S}\times\mathcal{A}_i\mapsto[0,1]$ parameterized by $\theta^i$, and produces the next state according to the state transition function $\mathcal{T}: \mathcal{S} \times \mathcal{A}_{1} \times \cdots \times \mathcal{A}_{N} \mapsto \mathcal{S}$.
Each agent $i$ obtains rewards based on function $r_{i}: \mathcal{S} \times \mathcal{A}_{i} \mapsto \mathbb{R}$. The initial states are determined by a distribution $\rho_0:\mathcal{S}\mapsto[0,1]$. Each agent $i$ aims to maximize its total discounted reward $R_i=\sum_{t=0}^{T}\gamma^t r_{i}^{t}$ where $\gamma$ is a discount factor and $T$ is the time horizon. In this paper, we consider a fully-cooperative Markov game where the reward is identical for all agents. To simplify notation, we omit $\theta$ from the subscript of $\pi$ when there is no ambiguity in the following sections.

\begin{figure*}[htb!]
	\centering
	\includegraphics[width=0.9\textwidth]{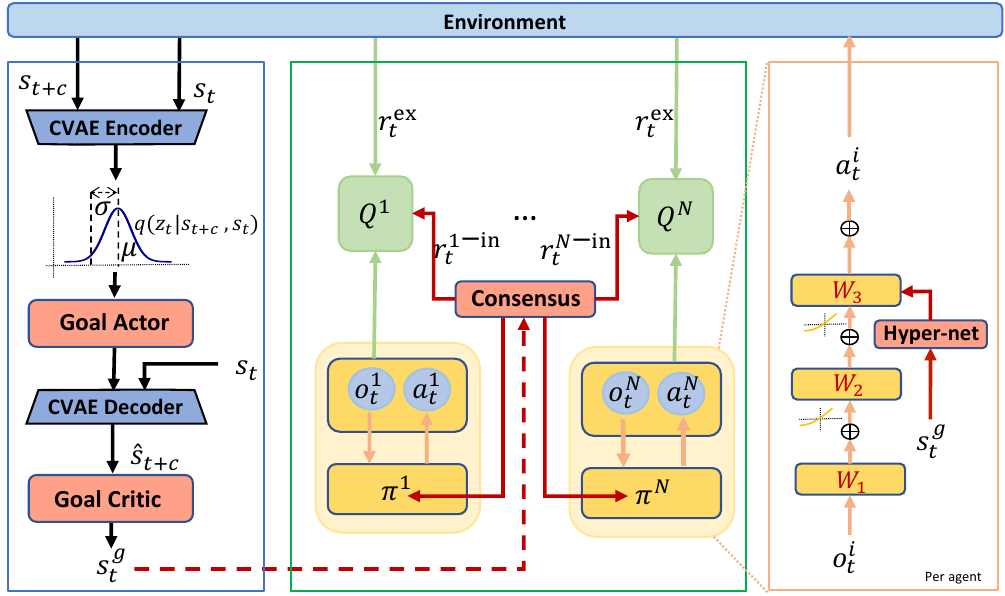}
	\text{(a) Goal imagination ~~~~~~~~~ (b) Multi-agent policy coordinated by goal consensus. ~~~~~ (c)Policy network.}
	\caption{Overview of MAGI.
	(a) The multi-agent goal imagination module. CVAE models the future state distribution, from which the goal actor and goal critic sample the common goal.
	(b) Agent network structure coordinated by imagined goal-based consensus mechanism with intrinsic reward and hypernetwork policy.
	(c) Policy network with goal-based hypernetwork.}\label{architecture}
\end{figure*}
\section{Methods} 
The proposed MAGI framework consists of a model-based goal imagination module and a model-free policy.
Figure~\ref{architecture} illustrates the overall architecture. 

In the next, we first introduce our model-based long-horizon goal imagination mechanism. We then explain how the generated goal helps agents reach consensus and coordinate model-free multi-agent learning. Finally, we present the whole training procedure of the proposed MAGI framework.

\subsection{Long-horizon Goal Imagination}
The long-horizon goal imagination module is a model-based structure that aims to establish a common goal as multi-agent consensus.
Specifically, this structure consists of two parts: a generative network for modeling future state distribution and an actor-critic-style \cite{konda2000actor} goal generator to produce a valuable long-horizon goal. 

\subsubsection{Future State Distribution Modeling}
We use CVAE to model the future state distribution, which is suitable for modeling long-term environment dynamics and easy for sampling possible states \cite{mishra2017prediction,krupnik2020multi}.
Furthermore, we decouple action selection from dynamics modeling, allowing our architecture to have better scalability in multi-agent environments.

Specifically, at time step $t$, we want to model a future state $s_{t+c}\in\mathcal{S}$ after a horizon length $c$. Our CVAE consists of a posterior distribution (encoder) $q_{\theta^{\text{enc}}}(z_t|s_{t+c},s_t)$, a generative distribution (decoder) $p_{\theta^{\text{dec}}}(s_{t+c}|s_t,z_t)$ and a prior distribution $p_{\theta^{\text{prior}}}(z_t|s_t)$ regularized by the posterior distribution $q_{\theta^{\text{enc}}}$.

As shown in Figure \ref{architecture}(a), the CVAE encoder maps $s_{t+c}$ and $s_t$ to a posterior distribution $q_{\theta^{\text{enc}}}(z_{t}|s_{t+c},s_t)$.
Then, we sample a $z_t$ from $q_{\theta^{\text{enc}}}(z_t | s_{t+c}, s_t)$ and feed it into decoder together with $s_t$ to model the distribution $p_{\theta^{\text{dec}}}(s_{t+c}|s_t,z_t)$.
The sampling process during training is differentiable using the reparameterization trick $z_t = \mu^{\text{post}} + \sigma^{\text{post}}\cdot\epsilon$, where  $\epsilon\sim\mathcal{N}(0,I)$ is the sample noise \cite{kingma2014auto}. 

\subsubsection{Goal Actor and Goal Critic}
After the future state distribution modeling, the goal state will be sampled and evaluated through an actor-critic-style process.   

The goal actor aims to find the hidden state from the prior distribution $p_{\theta^{\text{prior}}}(z_t|s_t)$ that can be decoded to the future state with the highest value.
We propose two types of goal actor with different sampling strategies:
\begin{itemize}
	\item	
	Uniform sampling \footnote{We use uniform sampling instead of widely used normal sampling to increase the probability of unexplored states.} $\epsilon$ within range $[-D,D]$, where D is a hyperparameter specifying sampling region.
	\item Training a deterministic sampling policy $\pi_{\psi}^{g}(s_t,\mu,\sigma)$ that outputs the reparameterization coefficient  $\epsilon$ directly. 
\end{itemize}

As a nonparametric method, the uniform sampling is more flexible, but each sample needs to be evaluated to compare values.
On the other hand, the parameterized deterministic policy $\pi_{\psi}^{g}(s_t,\mu,\sigma)$ has a lower computation cost during goal generation but requires training.
Our experimental results show that both strategies are effective.

The goal critic $V^g$ aims to evaluate the value of the imagined future state $\hat{s}_{t+c}$ which is decoded by the CVAE.
It can be trained using temporal-difference learning as the standard actor-critic method.


In summary, the general process of goal generation can be described as follows. 
\begin{itemize}
	\item For the uniformly sampling strategy, the CVAE decoder takes $M$ uniformly sampled hidden state [$z_t^1,z_t^2,\cdots,z_t^M$] as input and outputs a series of goal candidates [$\hat{s}_{t+c}^1,\hat{s}_{t+c}^2,\cdots,\hat{s}_{t+c}^M$]. Then, all goal candidates are fed into the goal critic to get the goal with highest state value $s_t^g=\arg\max_{\hat{s}_{t+c}^{i}}V^g(\hat{s}_{t+c}^{i})$, for $i \in \{1,2,\cdots, M\}$.
	\item For the deterministic sampling policy $\pi^g_{\psi}$, $\pi^g_{\psi}$ takes state $s_t,\mu$ and $\sigma$ as inputs and outputs a continuous action $\epsilon\in[-D,D]$. Then the hidden state $z_t=\mu+\sigma\cdot\epsilon$ is fed into the CVAE decoder to obtain the goal $s_{t}^g$.
\end{itemize}





\subsection{Consensus Coordinated Policy Learning}
After the long-horizon goal $s_t^g$ is generated, MAGI uses it to build consensus among agents.
MAGI provides two mechanisms for coordinating multi-agent policy learning with the goal-based consensus, as shown in Figure \ref{architecture}(b).


Firstly, the parameters of agents' policy network are generated by hypernetwork \cite{ha2016hypernetworks} taking the goal $s_t^g$ as input. 
This makes it possible to condition the weights of each agent's policy network on $s_t^{g}$ in arbitrary ways, thus integrating the goal imagination $s_t^g$ into each agent's policy as flexibly as possible. Then, we can perform standard actor-critic-based methods for policy learning. 

To further guide all agents to reach the consensus described by the goal state $s_t^{g}$, we design an intrinsic reward function that encourages agents to take actions leading to the goal state. 
Specifically, at time step $t$, we assign each agent an intrinsic reward based on whether the current action contributes to the goal state. Formally, the intrinsic reward for agent $i$ has the following form:
\begin{equation}
	r^{i\text{-in}}_{t} = d(u_i(s_t^g);v_i(s_{t})) - d(u_i(s_t^g);v_i(s_{t+1})),\label{eq:intrinsic_reward}
\end{equation}
$u_i(\cdot)$ and $v_i(\cdot)$ map $s_t^{g}$ and $s_t$ to the same metric space for agent $i$, and $d(\cdot)$ is the corresponding distance function.
Then, each agent $i$'s policy can be guided by a proxy reward function $r_{t}^{i\text{-proxy}}$ which combines the external environment's reward $r_{t}^{\text{ex}}$ and intrinsic reward $r^{\text{in}}_{t}$:
\begin{equation}
	r_{t}^{i\text{-proxy}} = r_{t}^{\text{ex}} + \lambda r_{t}^{i\text{-in}},\label{eq:new_reward}
\end{equation}
where $\lambda$ is a parameter balancing the external and intrinsic rewards. In the following training procedure, this proxy reward function $r_t^{i\text{-proxy}}$ is default reward setting for all agents.

\subsection{Training Procedure}
The long-horizon goal imagination module can be trained end-to-end together with the policy network. The training of CVAE is self-supervised, which takes state pairs $\{s_t, s_{t+c}\}$ from the agents' trajectory as training data.
Since the prior $p_{\theta^{\text{prior}}}(z_t|s_t)$ is regularized by $q_{\theta^{\text{enc}}}(z_t|s_{t+c},s_t)$, which is a Gaussian distribution with parameter $\mu$ and $\sigma$, the KL divergence in the evidence lower bound (ELBO) could be computed in closed form. Besides, the expected reconstruction error $\mathbb{E}_{q_{\theta^{\text{prior}}}(z_t|s_{t+c},s_t)}[\log p_{\theta^{\text{dec}}}(s_{t+c} | z_{t}, s_t)]$ can be estimated by sampling. 
Thus, maxing ELBO in CVAE is equivalent to minimizing the following loss function:
\begin{equation}
	\begin{aligned}
		\mathcal{L}_{\text{CVAE}}(\theta) &=  \mathbb{KL}[q_{\theta^{\text{enc}}}(z_t | s_{t+c}, s_t) \| p_{\theta^{\text{prior}}}(z_t|s_t)] \\&\quad- \mathbb{E}_{q_{\theta^{\text{enc}}}(z_t|s_{t+c},s_t)}[\log p_{\theta^{\text{dec}}}(s_{t+c} | z_{t}, s_t)].
	\end{aligned}\label{eq:cvaeloss}
\end{equation}

For the goal critic, it can be trained together with agent $i$'s centralized critic $Q^i$ for all $i\in\{1,2,\cdots,N\}$.
Formally, the goal critic loss has the following form:
\begin{equation}
	\mathcal{L}_{t}^{\text{gc}}(\zeta) = \frac{1}{N}\sum_{i=1}^{N}\left[V_{\zeta}^g(s_t)-Q^i(s_t,a_t^i)\right]^2.\label{eq:goal_critic_loss}
\end{equation}

Whether to train the goal actor depends on the sampling method of $z_t$.
The strategy of uniformly sampling $z_t$ in a neighborhood of mean vector $\mu$  does not require training since it is nonparametric.
For deterministic sampling policy $\pi_{\psi}^g$, we adopt a DDPG-style training and solve the following optimization problem using gradient ascent:
\begin{equation}
	\max_{\psi} V^g(f_{\text{dec}}(s_t, \mu+\sigma\cdot\pi_{\psi}^{g}(s_t,\mu,\sigma))),\label{eq:goal_actor_train}
\end{equation}
where the action of the goal actor $\pi_{\psi}^{g}$ is predicting the reparameterization coefficient $\epsilon$, and its objective is to maximize the value future state $s_t^{g}$ which is decoded by CVAE decoder $f_{\text{dec}}(z_t)$ with $z_t=\mu+\sigma\cdot\pi_{\psi}^{g}(s_t,\mu,\sigma)$. 
Specifically, we can use the following sampled policy gradient to optimize $(\ref{eq:goal_actor_train})$:
\begin{small}
\begin{equation}
	\nabla_{\psi}J(\psi)\approx\frac{1}{N_b}\sum_{j=1}^{N_b}\left[\nabla_{\psi}\pi_{\psi}^g(s_j,\mu,\sigma)\nabla_{\epsilon_j}V^g(f_{\text{dec}}(s_j,\mu+\sigma\cdot\epsilon_j))\right],\label{eq:goal_actor_loss}
\end{equation}
\end{small}
where $N_b$ is mini-batch size.


Each agent's policy optimization process follows the standard actor-critic paradigm such as DDPG and PPO \cite{schulman2017proximal}.
Figure\ref{architecture}(b) shows the policy optimization procedure using DDPG.
When using DDPG, agent $i$'s Q function can be optimized by minimizing the following loss:
\vspace{-1mm}
\begin{equation}
	\mathcal{L}(\theta^i) = \frac{1}{N_b}\sum_{j=1}^{N_b}\left(y_j^i-Q^i(s_j,a^i_j;\theta^i)\right)^2,\label{eq:policy_critic_loss}
\end{equation}
\vspace{-1mm}
where $y_j^i=r_j^{i\text{-proxy}}+\gamma Q^{'i}({s'_j},{a'_j}^{i};{\theta^{'}}^{i})|_{{a'_j}^{ i}=\pi^i({s'_j})}$ is the one step temporal difference target, $N_b$ is the batch size, and $Q^{'i}$ is the target network.
Each agent $i$'s policy $\pi^i$ can be updated with the sampled policy gradient:
\begin{small}
	\begin{equation}
		\begin{aligned}
			\nabla_{\phi^i}J(\phi^i)\approx \frac{1}{N_b}\sum_{j=1}^{N_b}\nabla_{\phi^i}\pi^{i}(s_j;\phi^i)\nabla_{a_j^i}Q^{i}(s_j,a^{i}_j),\label{eq:policy_actor_loss}
		\end{aligned}
	\end{equation}
\end{small}

It is worth noting that the DDPG structure mentioned above for agent's policy training can be replaced by other actor-critic-based algorithms, such as PPO \cite{schulman2017proximal}. See the experiment section for more details.


\section{Experiments}
We evaluate MAGI on two multi-agent environments: the multi-agent particle-environments (MPEs) and challenging Google Research Football (GRF). In this section, we present experimental results and answer the following three questions:
1) Can MAGI achieve better performance on MPEs and challenging GRF envrionments?
2) How does the consensus modeling improve the performance?
3) Does the generated goal actually influence the agents' policy?

\subsection{Environments}
\begin{figure}
    \centering
	\subfigure[Navigation]{
		\label{fig:navigation-scene}
		\includegraphics[width=38mm]{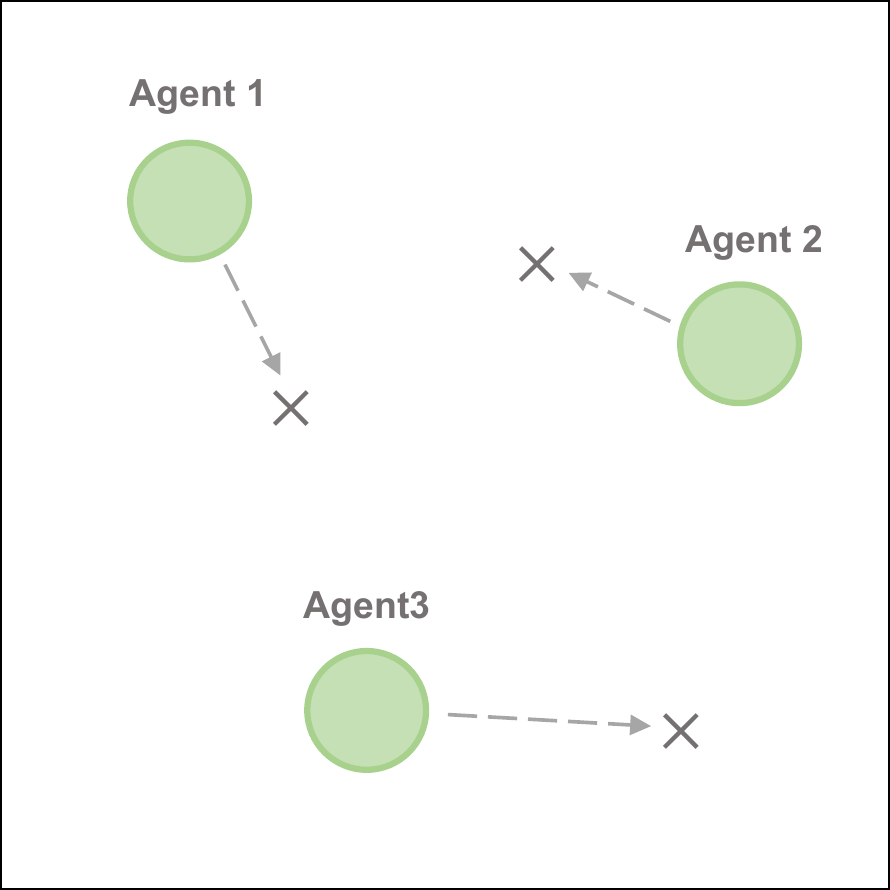}
	} \quad
	\vspace{-2mm}
	\subfigure[Treasure Collection]{
		\label{fig:collect-scene}
		\includegraphics[width=38mm]{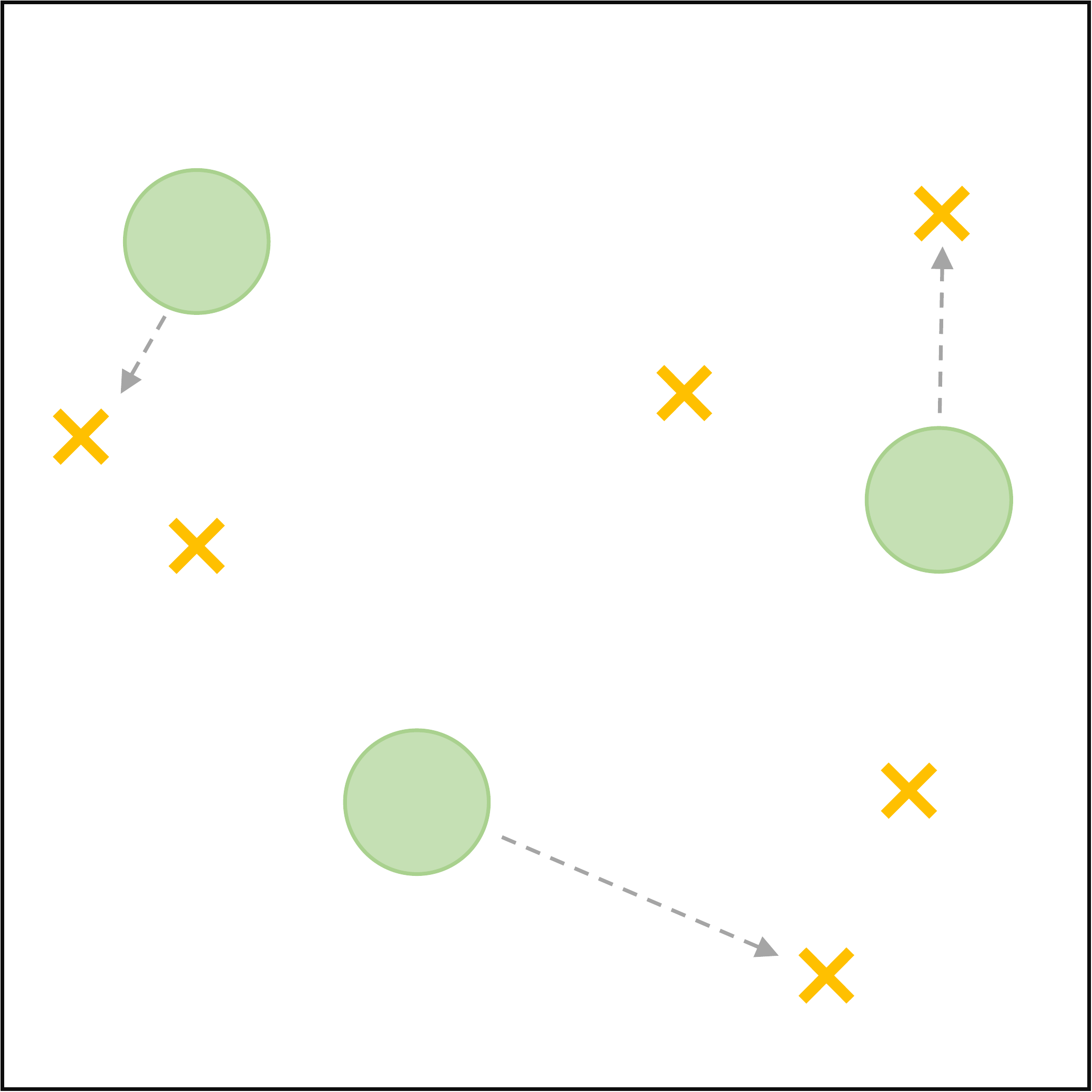}
	}
	\vspace{-3mm}
	\subfigure[Predator-Prey]{
		\label{fig:predator-scene}
		\includegraphics[width=38mm]{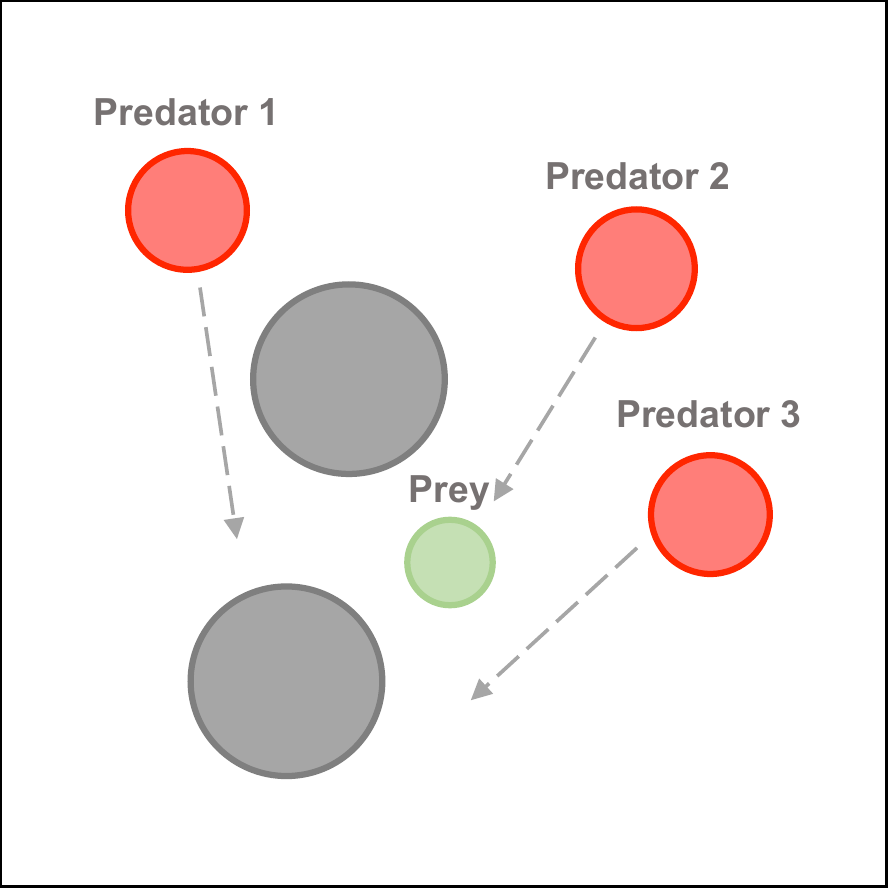}
	} \quad
	\subfigure[Keep-Away]{
		\label{fig:keepaway-scene}
		\includegraphics[width=38mm]{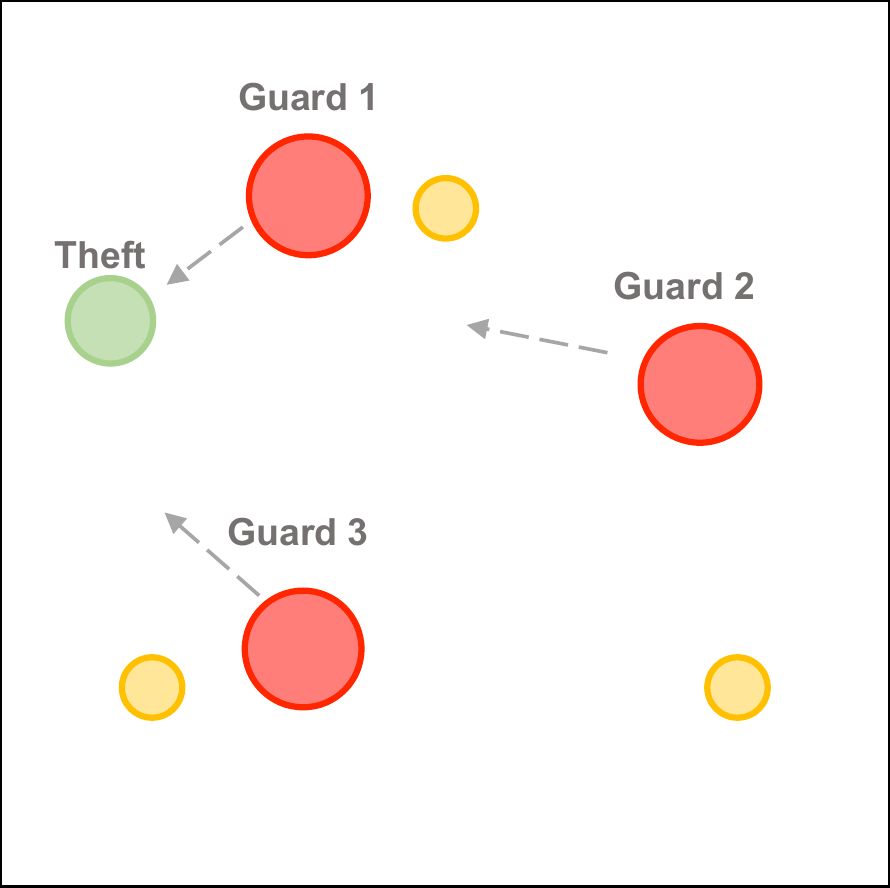}
	}
	\caption{Illustrations of the Multi-agent Particle-Environments.} 
	\label{fig:MPE-scenes}
\end{figure}

\begin{figure*}
	\centering
	\hspace{-2mm}
	\vspace{-3mm}
	\subfigure[Navigation]{
		\label{fig:navigation}
		\includegraphics[width=43mm]{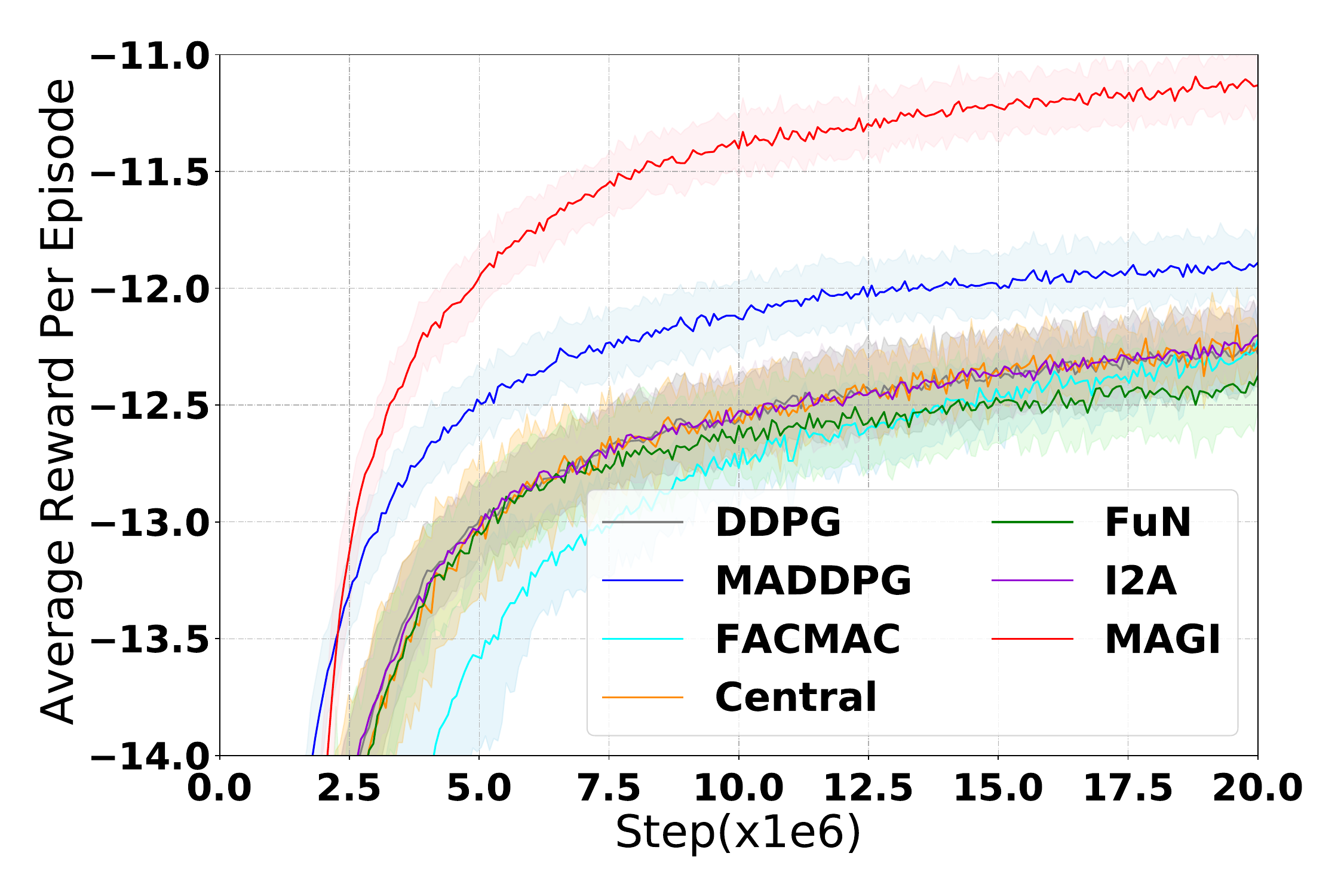}
	}
	\hspace{-2mm}
	\subfigure[Treasure Collection]{
		\label{fig:deception}
		\includegraphics[width=43mm]{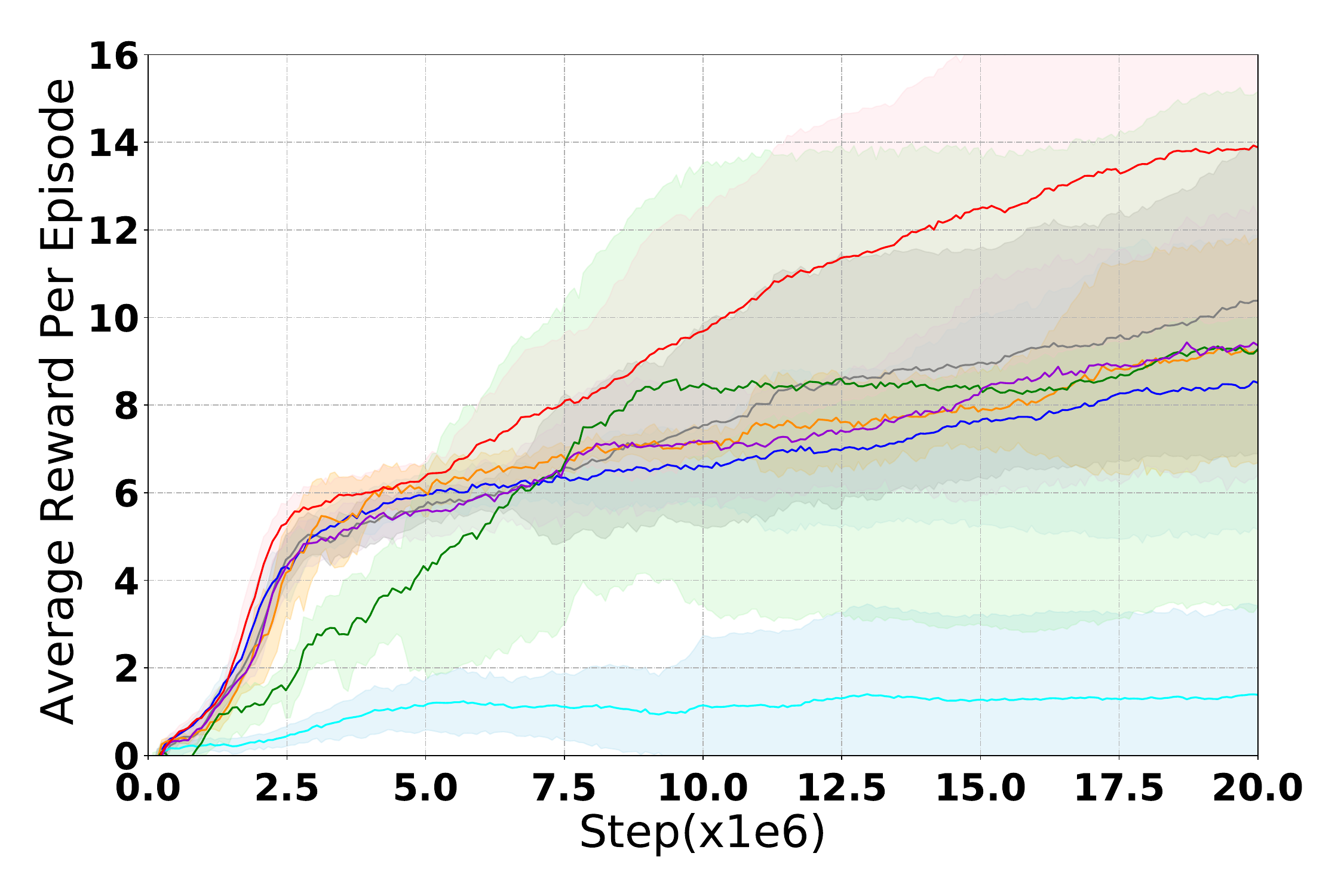}
	}
	\hspace{-2mm}
	\subfigure[Predator-Prey]{
		\label{fig:predator}
		\includegraphics[width=43mm]{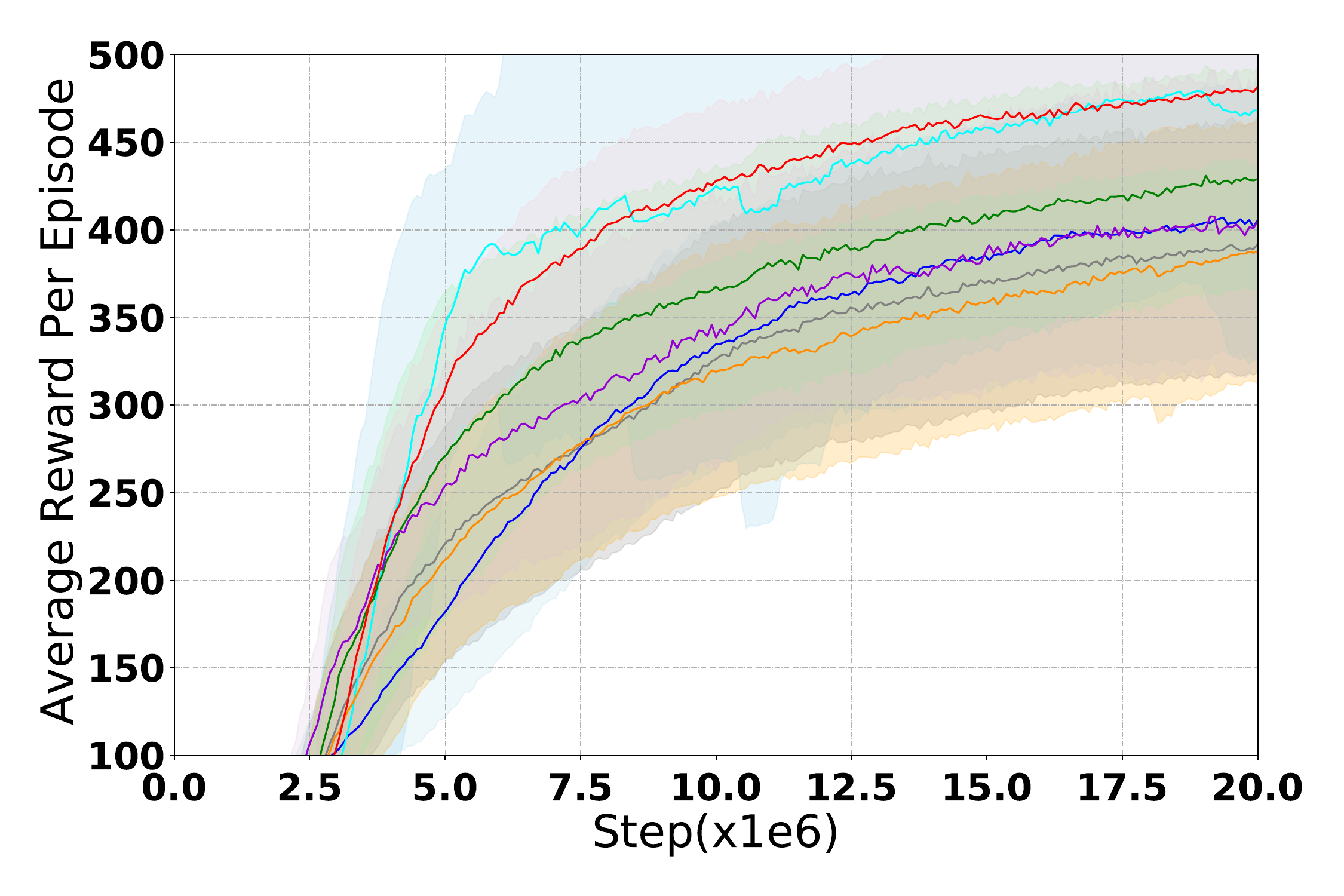}
	}
	\hspace{-3mm}
	\subfigure[Keep-Away]{
		\label{fig:keepaway}
		\includegraphics[width=43mm]{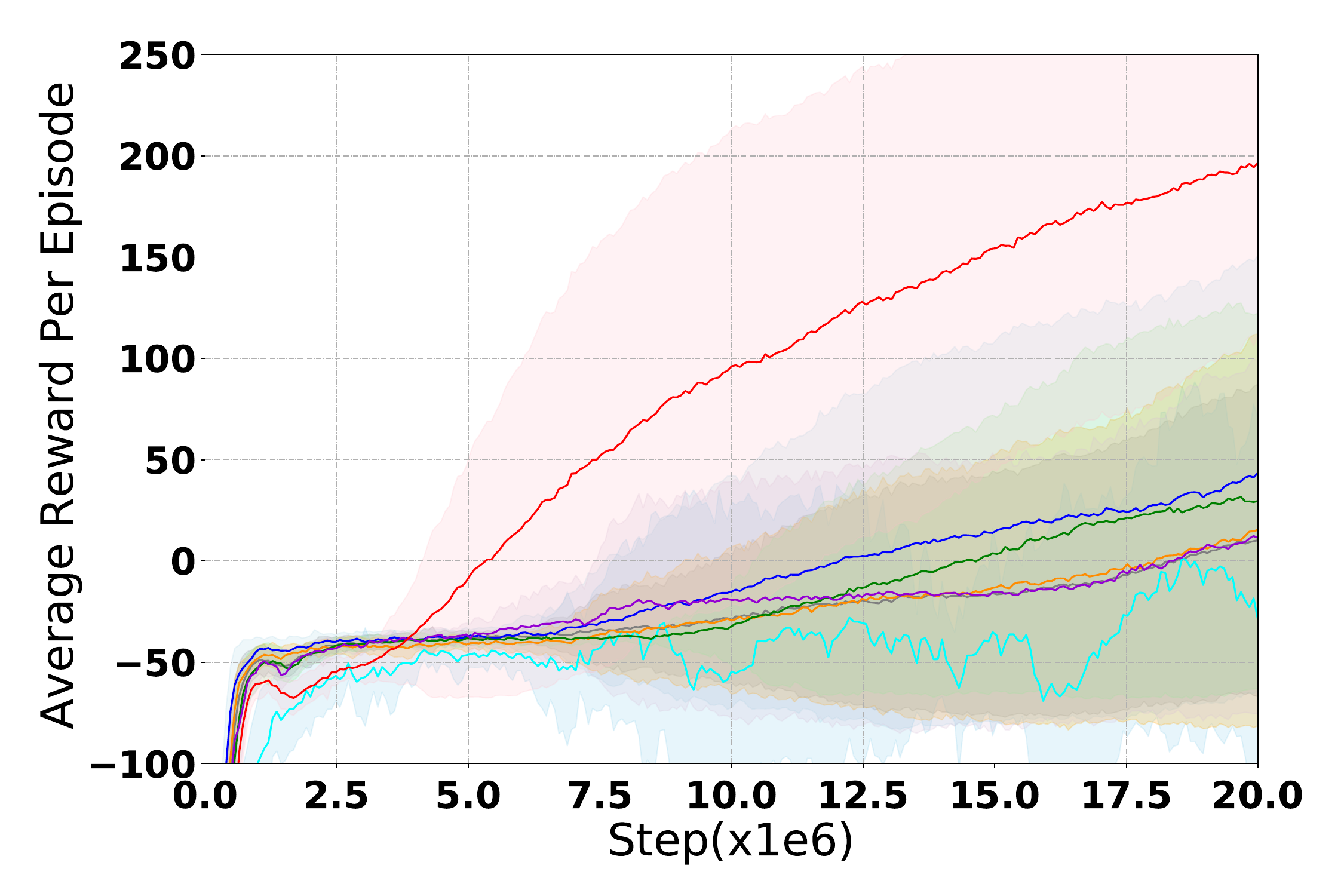}
	}
	\caption{The average results in four MPEs tasks. MAGI outperforms all other methods in sample efficiency and performance. The legend in (a) applies across all plots.} 
	\label{fig:MPEs}
\end{figure*}

\subsubsection{Multi-agent Particle-Environments}
The MPEs is a physics-based 2D environment, which provides continuous action space, including stay and move. We consider the following four tasks as shown in Figure \ref{fig:MPE-scenes}:

\textit{Navigation} $N=3$ agents need to cooperatively cover 3 landmarks while avoiding collision.

\textit{Treasure Collection} \cite{iqbal2019actor} $N=3$ agents aim to cooperatively collect 6 randomly placed treasures. 

\textit{Ten-Agents Treasure Collection} $N=10$ agents to collect 20 treasures. This is an extension of Treasure Collection.

\textit{Predator-Prey} $N=3$ slower predators aim to catch one faster prey which is controlled by a fixed pretrained policy.

\textit{Keep Away} $N=3$ slower guards agents need to safeguard three treasures.
Another faster theft agent controlled by a fixed pretrained policy tries to steal treasures. 

In Predator-Prey and Keep Away tasks, the fixed opponent agent is pretrained by self-play using DDPG. For detailed reward designs, please refer to Appendix B.

\subsubsection{Google Research Football}
The GRF is a challenging physics-based 3D football simulation environment.
It provides large discrete action space, including moving, shooting, and passing.
Agents need complex cooperation to get more scores than opponents.
In \textit{N vs N} scenario, there are $N-1$ players and one goalkeeper on each team.
We consider scenarios \textit{3 vs 3},  \textit{4 vs 4} and  \textit{5 vs 5}, in which the opponent team is controlled by built-in AI.


\subsection{Baselines and Settings}
For MPEs, we build MAGI based on DDPG \cite{Lillicrap2016}.
We compare MAGI with the following approaches:
centralized DDPG;
CTDE methods MADDPG \cite{lowe2017multi} and FACMAC \cite{peng2021facmac} (an extension of QMIX under actor-critic framework);
model-based planning RL methods Conditional-TSM \cite{krupnik2020multi} and MPC \cite{nagabandi2018neural};
an imagination-augmented method I2A \cite{racaniere2017imagination};
a goal-conditioned hierarchical method Feudal Network (FuN) \cite{vezhnevets2017feudal}. 
The implementations of DDPG, MADDPG, FACMAC, I2A, and FuN adopt the same policy network structure as MAGI.
During the evaluation of model-based planning methods, we use the same planning horizon length as MAGI and only choose the Navigation task considering the heavy computation complexity of learning dynamics model in other scenarios. We list the comparison of parameter and training complexity in Table \ref{tab:complexity_compare}.

\begin{table}[ht]
\centering
\setlength{\tabcolsep}{4mm}
\begin{tabular}{cccc}
\toprule
       Methods          &  \thead{Parameter\\Number} & \thead{Inference\\Time} & \thead{Training\\Time}   \\  
\midrule
DDPG          &   1 $\times$  & 1 $\times$ & 1 $\times$ \\
MADDPG          &   1 $\times$  & 1 $\times$ & 1.05 $\times$ \\
FACMAC          &   1 $\times$  & 1 $\times$ & 0.78 $\times$ \\
Central          &   1.03 $\times$  & 0.47 $\times$ & 0.50 $\times$ \\
FuN                     & 1.59 $\times$ & 1.23 $\times$  & 1.40 $\times$   \\
I2A                     & 4.42 $\times$ & 285.70 $\times$  & 15.23 $\times$   \\ 
\textbf{MAGI }           &   \textbf{1.54} $\times$ & \textbf{1.22} $\times$ & \textbf{1.54} $\times$  \\  
\bottomrule
\end{tabular}
\caption{Parameter number, inference and training time comparison of different methods (DDPG is normalized to 1x).}
\label{tab:complexity_compare}
\end{table}

For the Google Research Football environment, we choose PPO as the baseline since its performance in complex environments has been verified before \cite{berner2019dota,ye2020towards}. We build MAGI upon PPO to verify its effectiveness in complex environments. 2 Nvidia V100 GPUs and 800 AMD EPYC 7K62 CPU cores are used for GRF experiments.

The imagination module takes $c = 4$ and $c = 16$ as the horizon length in MPEs and GRF.
Unless specified, we use uniform sampling in experiments.
When implementing intrinsic reward in Equation \ref{eq:intrinsic_reward}, $u_i$ and $v_i$ output the coordinates of the agent $i$'s goal position and current position, $d$ is set as Euclidean distance, and the goal reward weight is set as $\lambda = 0.001$.
For environments where the distance is hard to measure directly, we provide a generalized implementation of intrinsic reward. Please refer to Appendix A for details.



\subsection{Main Results}
For tasks in MPEs, we repeated each experiment 32 times, and the average results are displayed in Figure \ref{fig:MPEs} and Table \ref{tab:converge_res}.
Results show that \MN outperforms all baselines in every task regarding sample efficiency and final performance.

Compared with baseline methods DDPG and Centralized-DDPG, \MN achieves better performance and sample efficiency.
This is likely because the common goal in MAGI improves agents' exploration and coordination.

MAGI yields better and more stable results than MADDPG and FACMAC, two representatives of CTDE methods that implicitly take consensus into consideration during training. This result demonstrates the importance of an explicit consensus mechanism for multi-agent coordination.

\begin{table}[bt]
\centering
\setlength{\tabcolsep}{7mm}
\begin{tabular}{cc}
\toprule
       Methods          &  Rewards  \\  
\midrule
Conditional-TSM          &   -20.3     \\  
MPC                      &   -17.6     \\ 
\textbf{MAGI }           &   \textbf{
-11.1 }    \\  
\bottomrule
\end{tabular}
\caption{Results of Conditional-TSM and MPC in Navigation task.}
\label{tab:converge_res}
\end{table}

\begin{figure*}[ht]
    \centering
    \subfigure[GRF environment]{
        \label{fig:gfootball_scenario}
        \includegraphics[width=40mm]{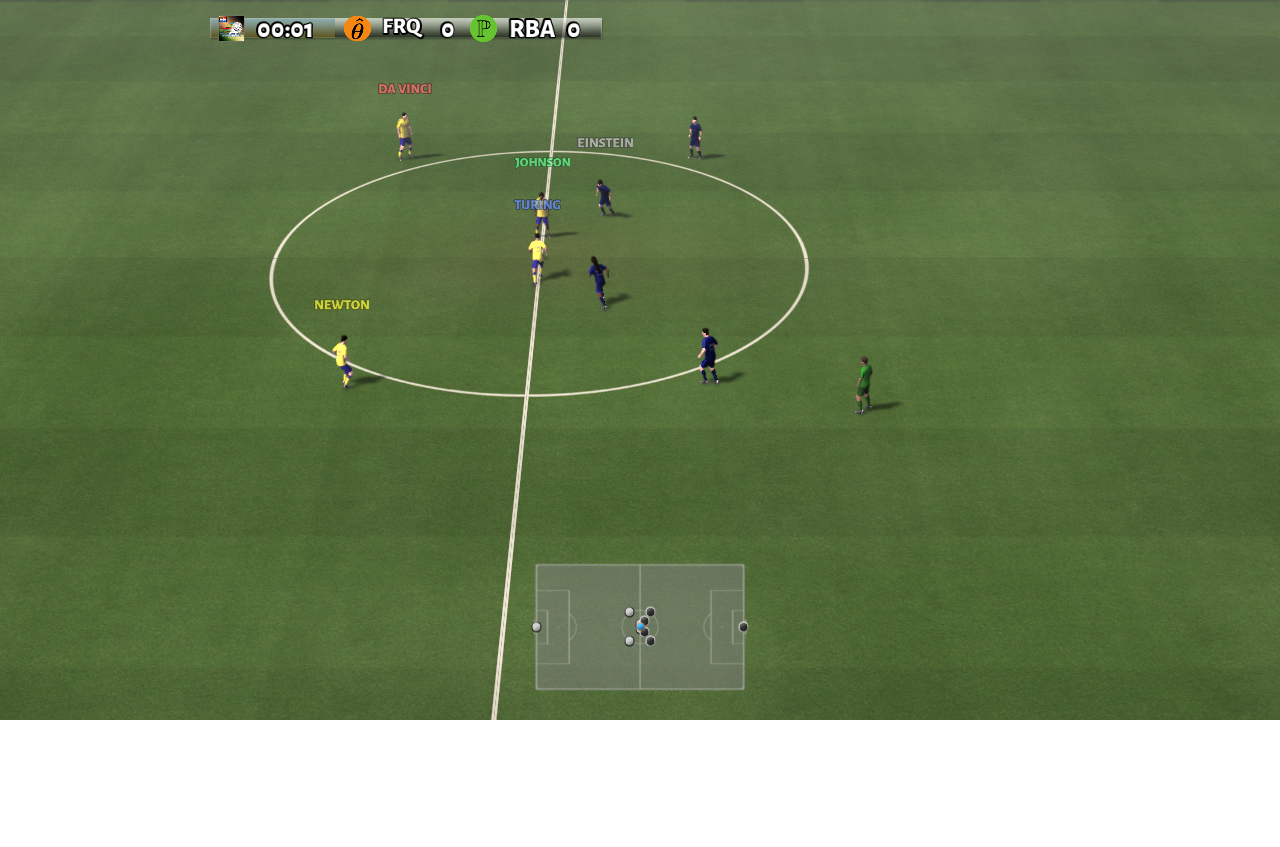}
    }
    \subfigure[3 vs 3]{
        \label{fig:football_3v3}
        \includegraphics[width=42mm]{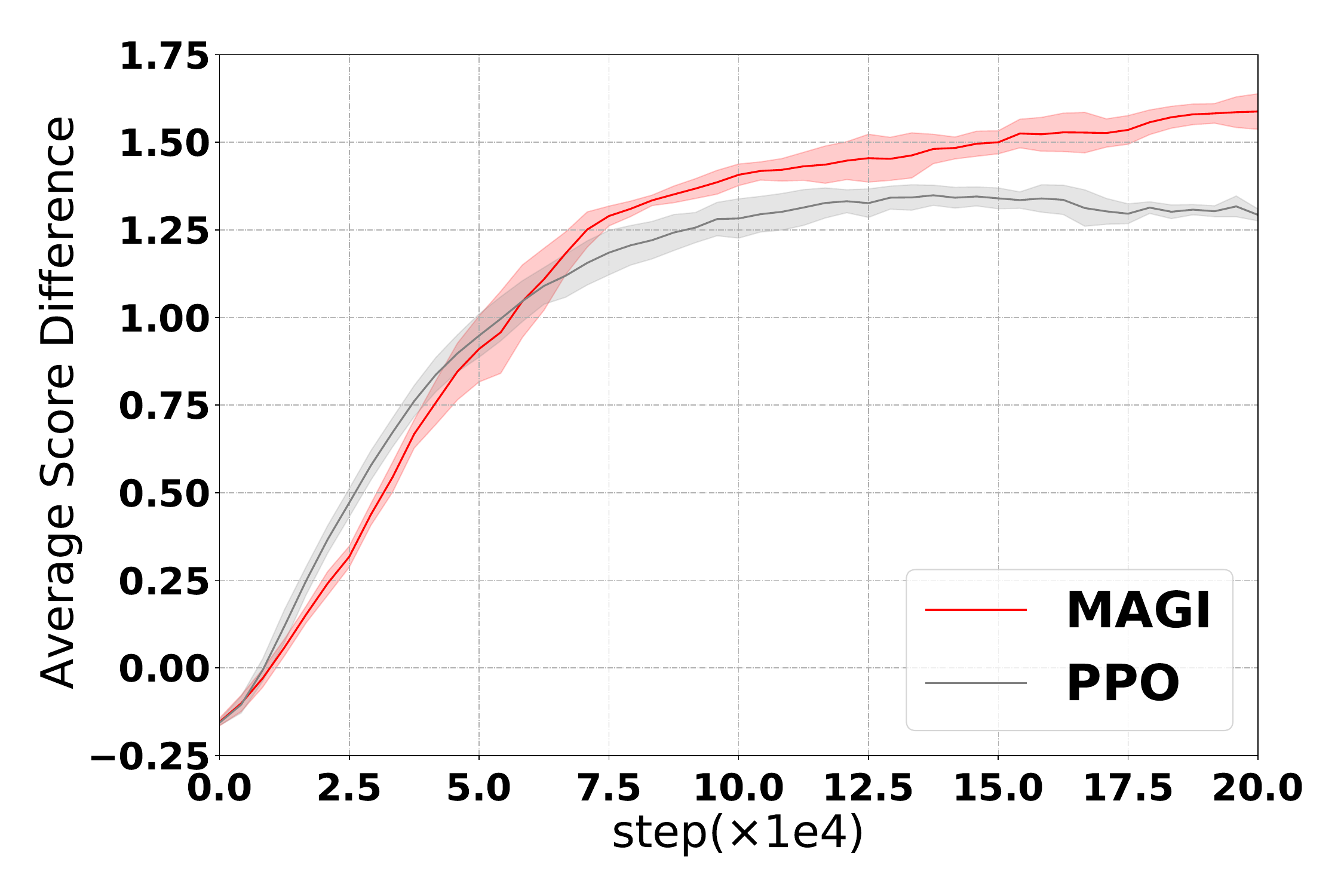}
    }
    \subfigure[4 vs 4]{
        \label{fig:football_4v4}
        \includegraphics[width=42mm]{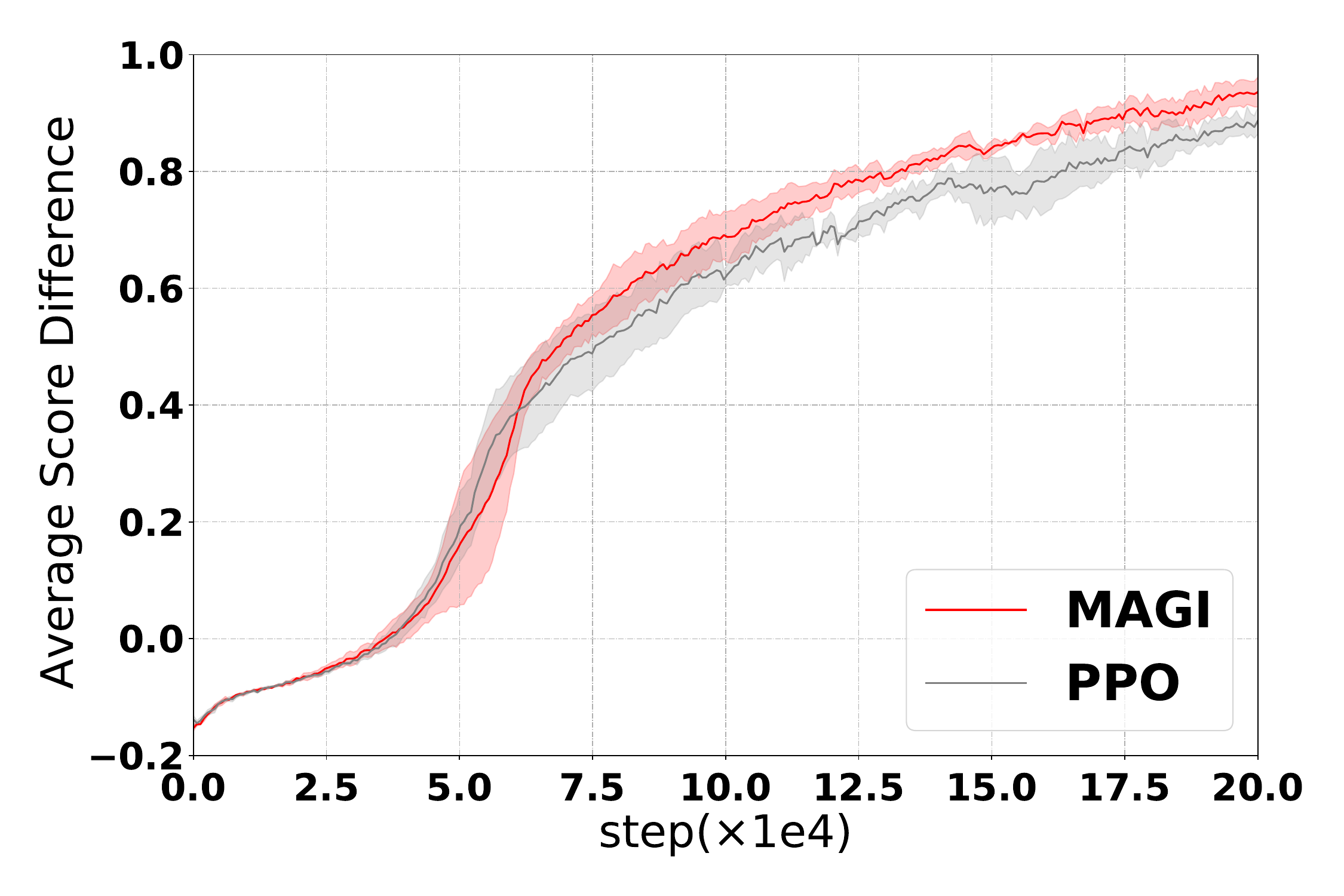}
    }
    \subfigure[5 vs 5]{
        \label{fig:football_5v5}
        \includegraphics[width=42mm]{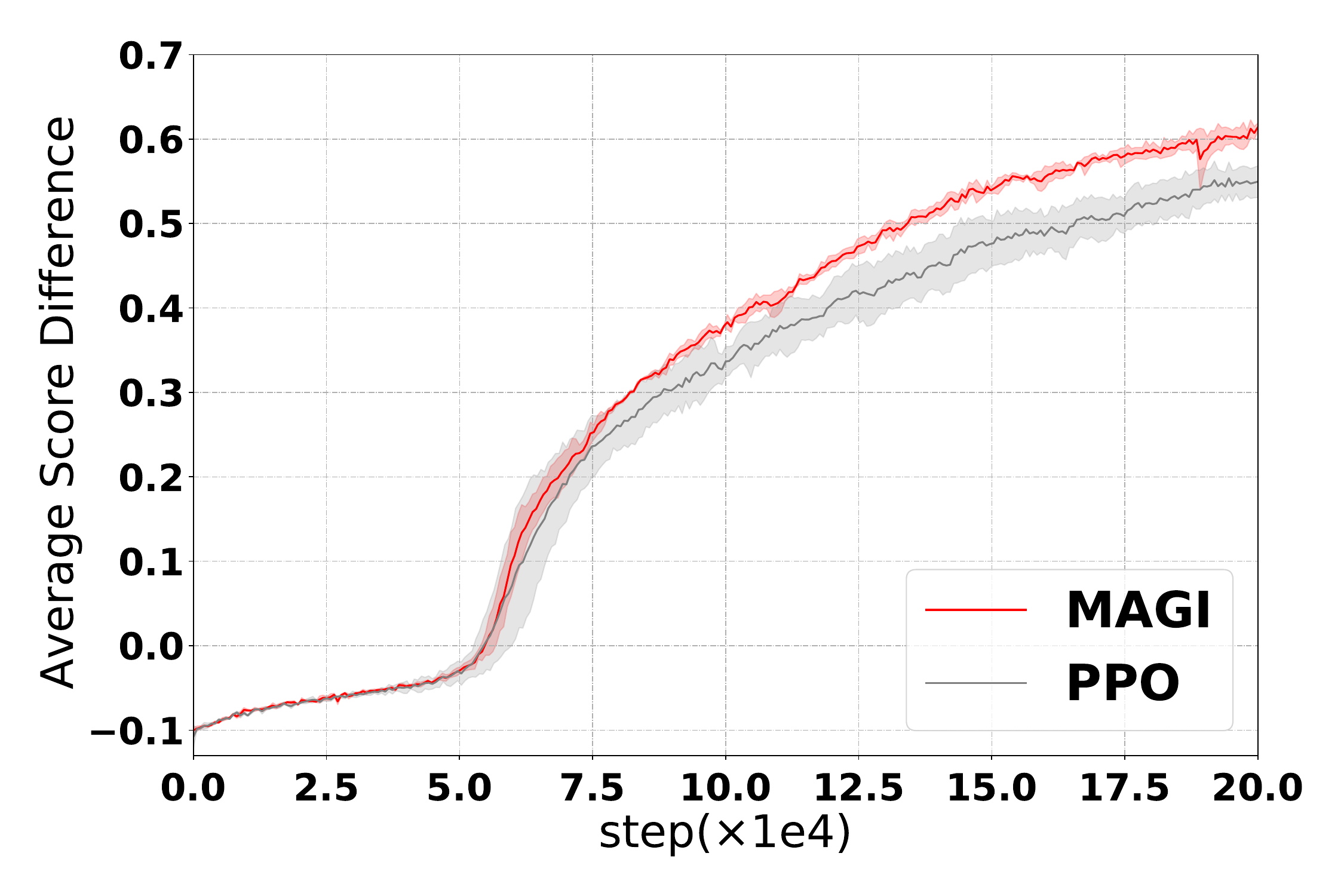}
    }
    \caption{Demonstration of the GRF environment and average goal difference results in three GRF scenarios.}
    \label{fig:GRF}
\end{figure*}

MAGI significantly outperforms model-based methods Conditional-TSM and MPC.
Its performance can likely be attributed to that MAGI directly models long-term dynamics, which decouples high-level state planning and low-level action decision. Therefore it can suffer less from dynamics prediction compounding error and produce better goals to guide agents cooperation. 

\MN performs better than the goal-conditioned method FuN, which uses hidden state as the goal. We assume that the goal generated by self-supervised generative method in \MN can provide more direct and useful instructions for agents than the hidden state goal of FuN.

Compared with I2A, an imagination-augmented method reaches significantly better performance. The result can be attributed to that our imagination module can generate better future states and consensus mechanism can coordinate agents more efficiently.

Average results in GRF are shown in Figure \ref{fig:GRF}.
MAGI yields better results than PPO in all scenarios, including the challenging \textit{5 vs 5}, showing that MAGI has the ability to solve complex multi-agent tasks and has the adaptability to be built upon different learning algorithms.

\begin{figure}[ht]
	\centering
	\includegraphics[width=60mm]{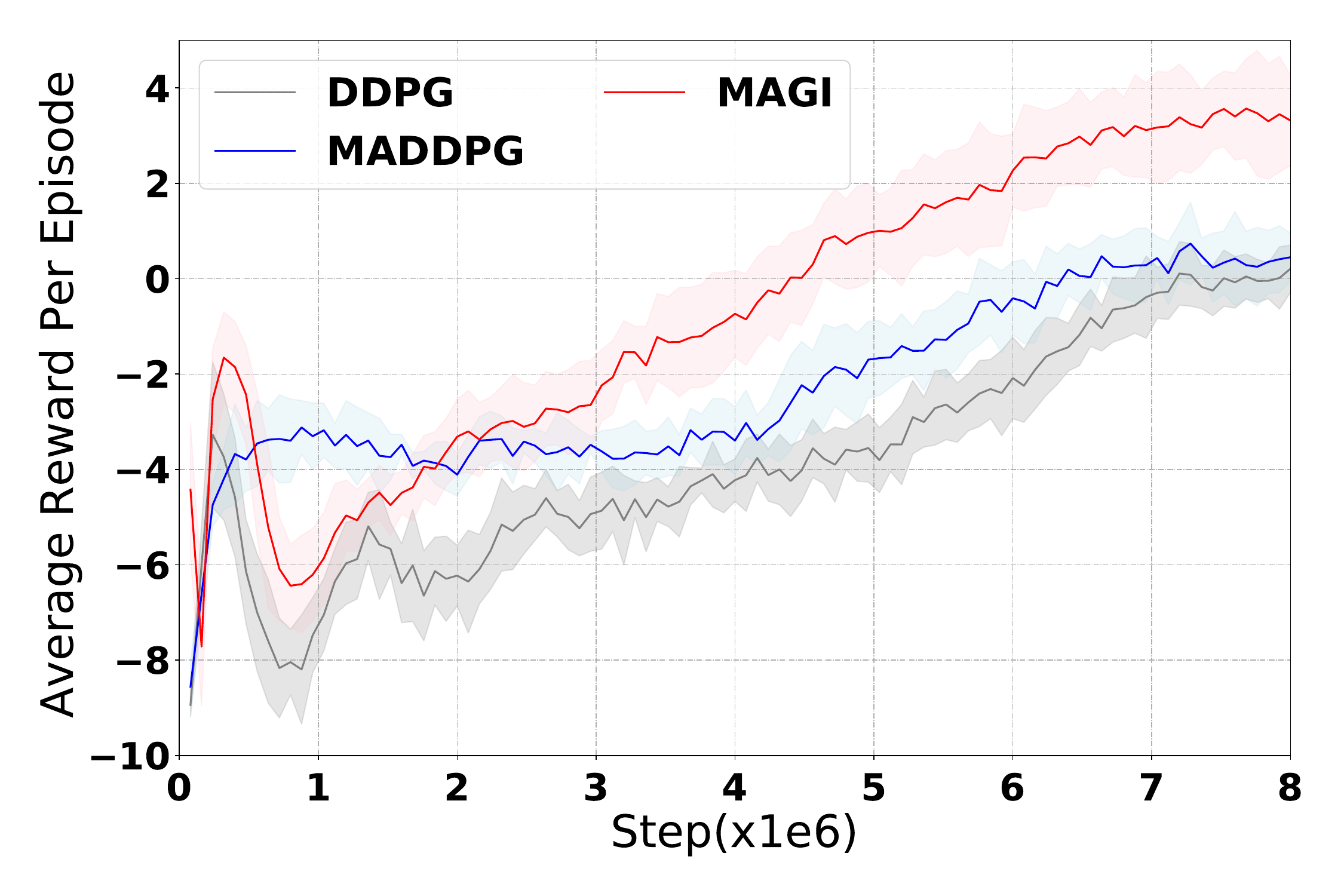}
	\caption{Demonstration of scalability. MAGI improves performance in Ten-Agents Treasure Collection task.}
	\label{fig:scalability}
\end{figure}

To further demonstrate the scalability of MAGI, we expand agent number to 10 in Treasure Collection task and the results are shown in  
 Figure \ref{fig:scalability}.
MAGI achieves better performance in comparison of baseline DDPG and MADDPG methods.
This result shows that MAGI can be scaled to scenarios with large number of agents.

\begin{figure}[t]
	\centering
	\vspace{-2mm}
 	\subfigure[Step 1M]{
		\label{fig:states_visual_0)}
		\includegraphics[width=38mm]{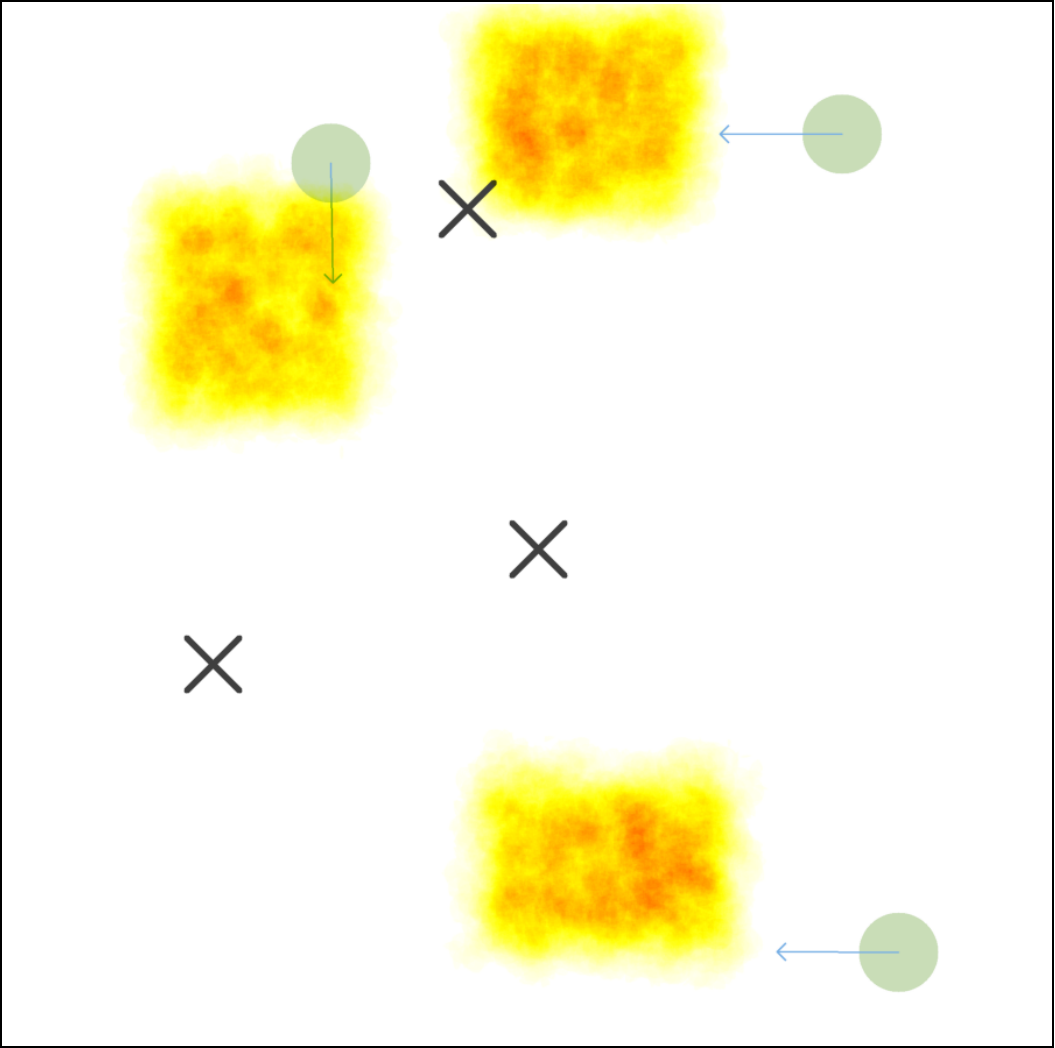}
	}
        \hspace{2mm}
	\subfigure[Step 20M]{
		\label{fig:states_visual_1}
		\includegraphics[width=38mm]{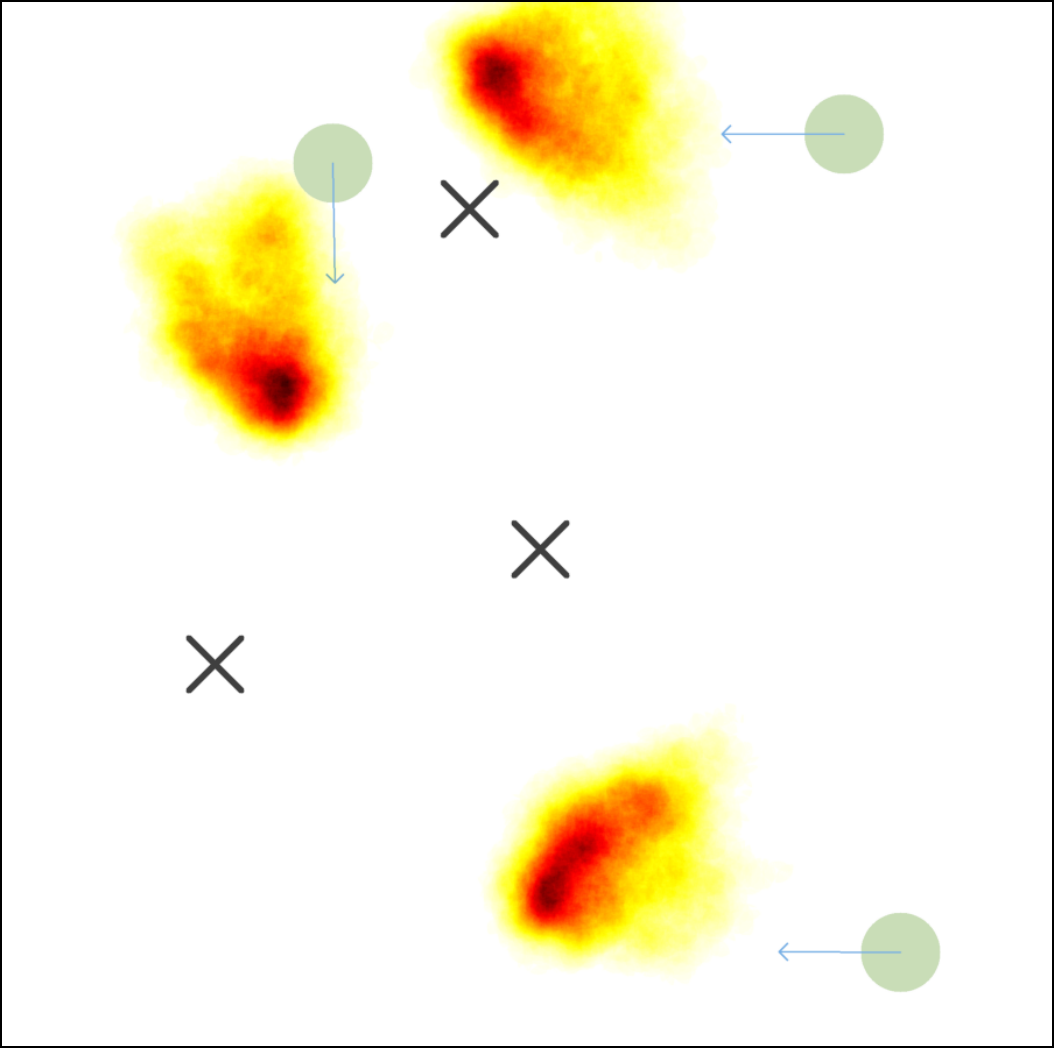}
	}
        \hspace{-2mm}
	\subfigure[Step 1M]{
		\label{fig:traj_visual_0)}
		\includegraphics[width=38mm]{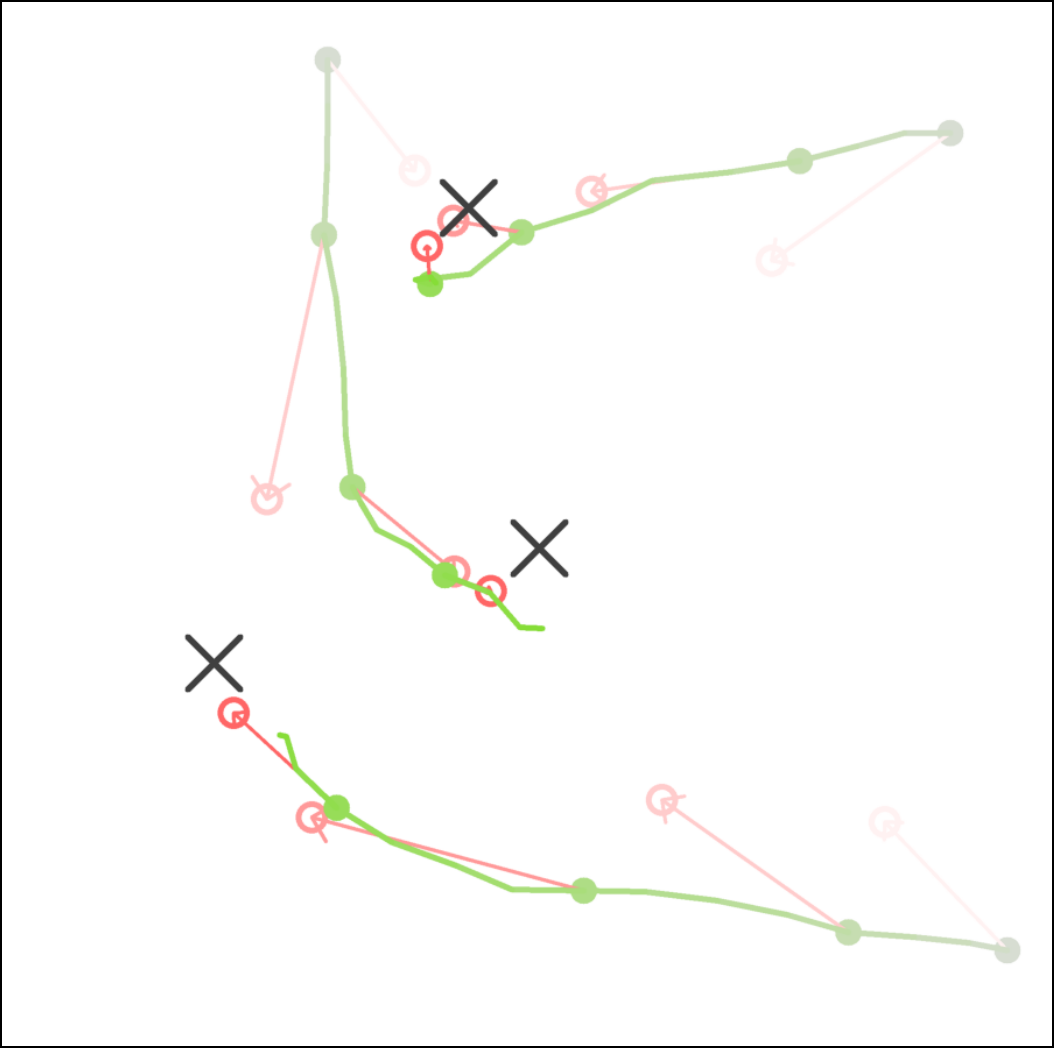}
	}
        \hspace{2mm}
	\subfigure[Step 20M]{
		\label{fig:traj_visual_1}
		\includegraphics[width=38mm]{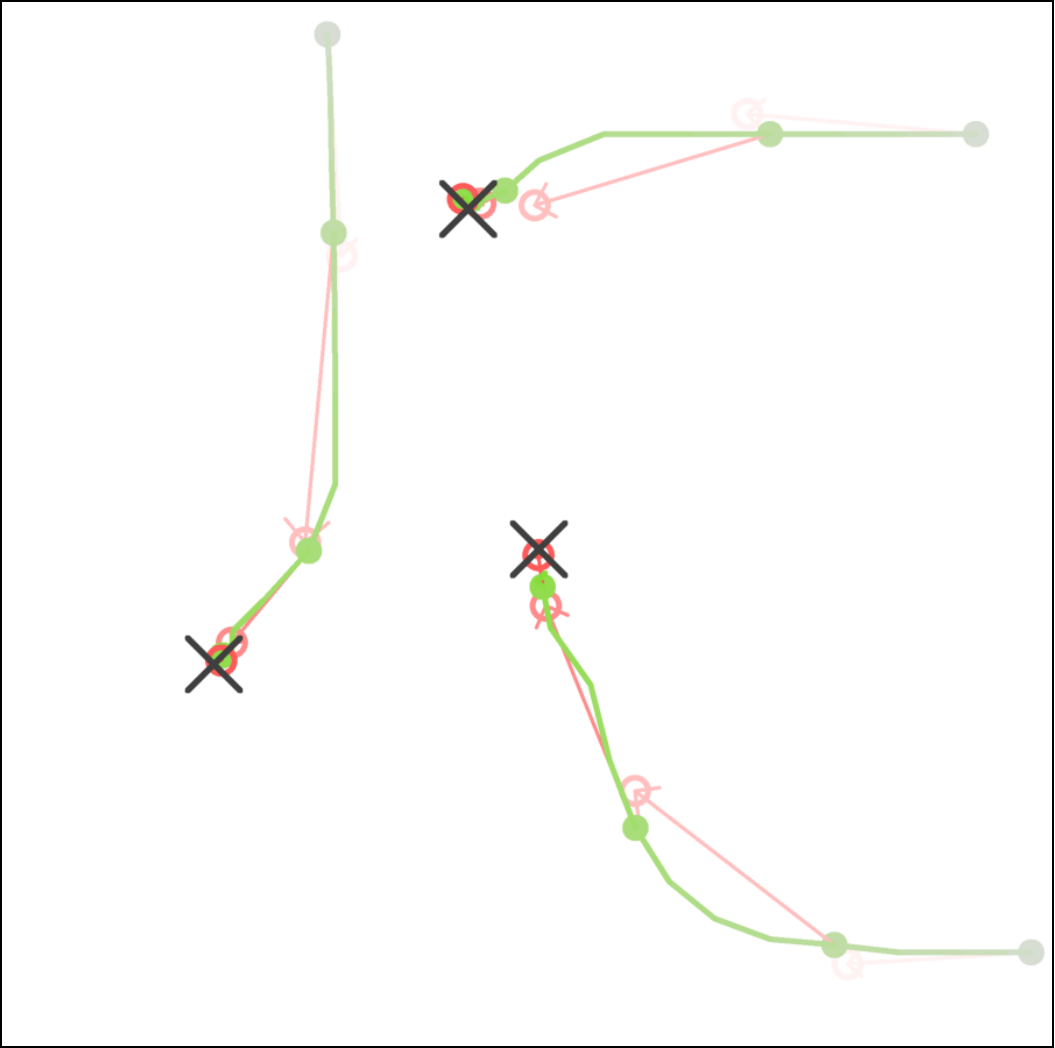}
	}
	\caption{(a) and (b): distribution of imagined positions of agents. Green circles and blue arrows are agents' current positions and velocities, and black crosses are target landmarks. (c) and (d): illustration of trajectories and imagined goals. The green curves are trajectories of agents, and red circles are positions of goals.}
	\label{fig:states_visual}
\end{figure}

\subsection{Consensus Analysis}
To verify how the learned consensus in \MN affect policies of agents, we conduct several visualization analyses.

The visualization of the imagined future state distributions are presented in Figure \ref{fig:states_visual}(a) and (b).
Imagined states at training step 20M are concentrated in regions closer to target landmarks than training step 1M, demonstrating that MAGI can provide better imagination about future states after training, which helps to sample valuable goals more efficiently. From multi-agent coordination perspective, the difference in the distribution of each agent's future goal makes it possible for all agents as a team to get more rewards, thus achieving consensus among all agents.

Another question is whether agents can perform actions according to the common goal (or team consensus). We further provide a case study to visualize the agents' trajectories and goals in Figure \ref{fig:states_visual}(c) and (d).
At training step 1M, goals are far away from targets, and agents fail to reach the goals.
When the training step comes to 20M, goals can provide more useful instructions, and agents follow the goals well.

The above analysis indicates that MAGI can learn to generate achievable and valuable common goals for all agents, which achieves a consensus mechanism and further improves multi-agent coordination.

\subsection{Ablation Study} \label{subsec:goal_effectiveness}
We conducted two ablation experiments in Navigation task to evaluate the effectiveness of the goal generated by MAGI.

\begin{figure}[ht]
	\centering
	\hspace{-5mm}
	\subfigure[Sample Size]{
		\label{fig:sample-num)}
		\includegraphics[width=42mm]{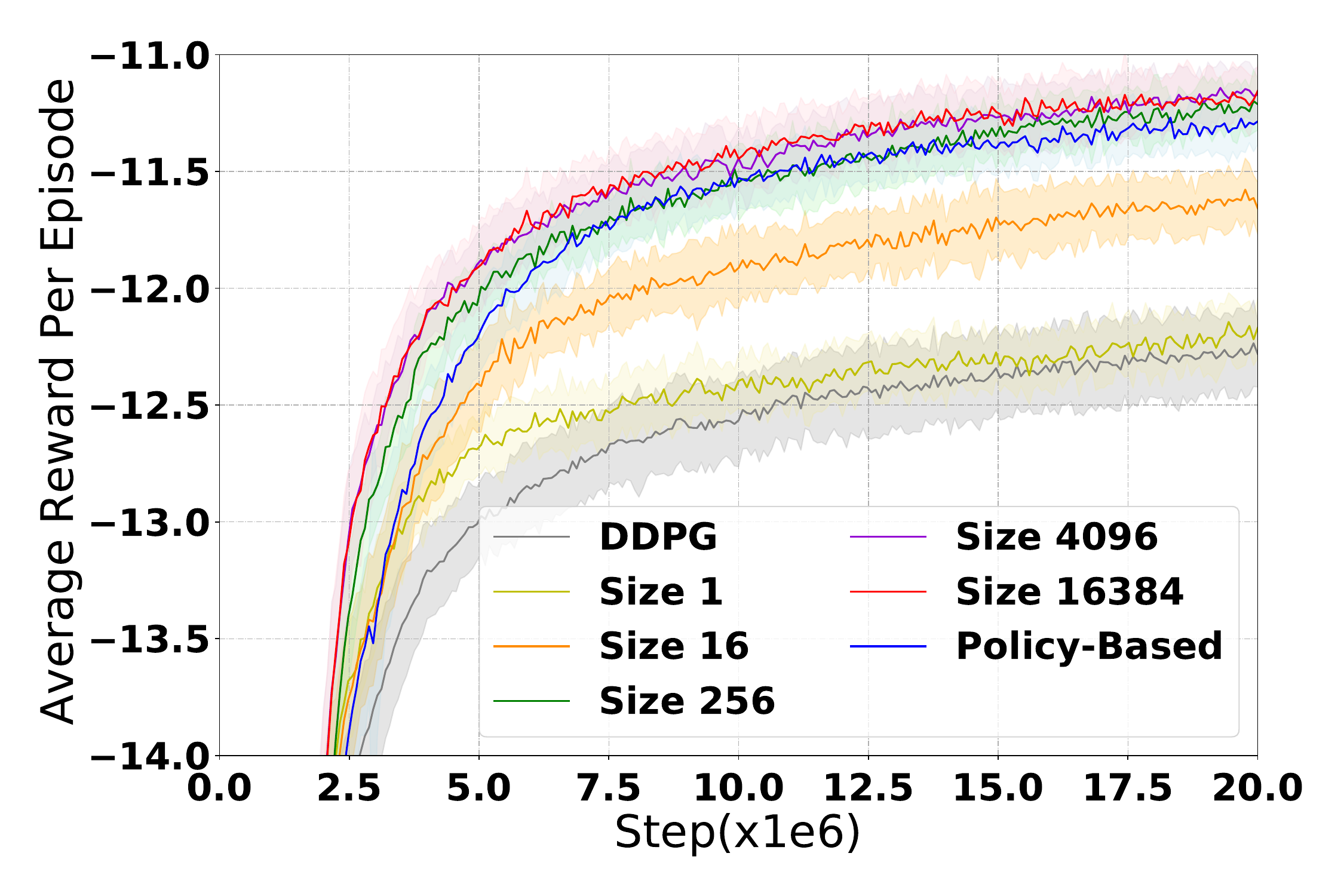}
	} 
	\hspace{-5mm}
	\subfigure[Horizon Length]{
		\label{fig:horizon}
		\includegraphics[width=42mm]{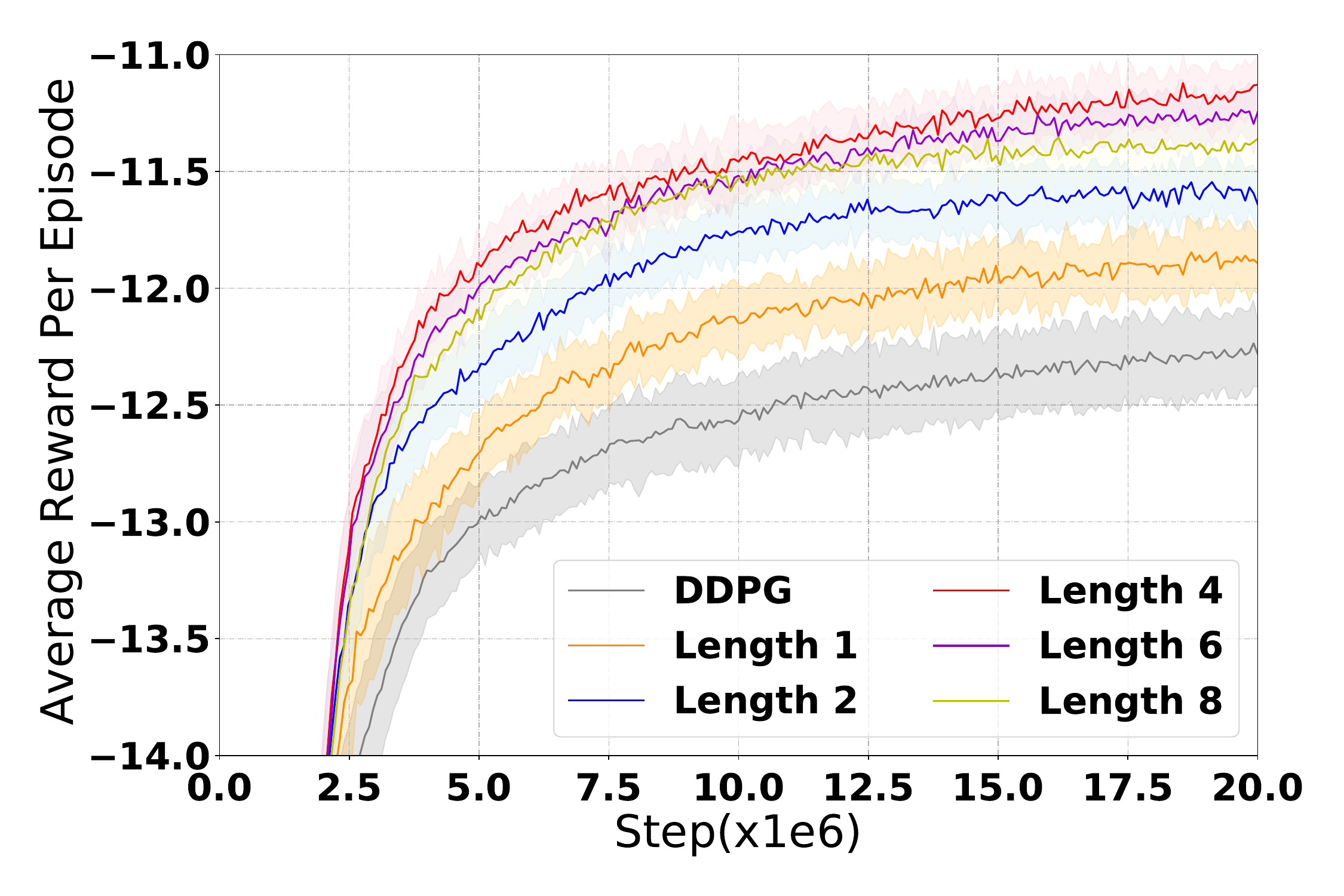}
	}
	\caption{Ablation results of sample size and horizon length.}
	\label{fig:effectiveness_magi}
\end{figure}

We first analyze the influence of sample size and sampling strategy.
Figure \ref{fig:sample-num)} shows that when the sample size is 1, MAGI degrades to naive DDPG.
As the sample size increases, the quality of generated goals becomes better, and the performance of MAGI improves monotonically.
The deterministic sampling also achieves comparable performance with the uniformly sampling strategy.

Then, we test MAGI with different planning horizon lengths.
As shown in Figure \ref{fig:horizon}, horizon length that is too short or too long both degrade the performance.
We attribute it to that short horizon length goal provide little useful instructions. While excessive long horizon goals are hard for agents to achieve, thus failing to guide agents effectively.

These results prove the goal generated by MAGI actually coordinates the policies of agents and a valuable and achievable goal is of vital importance for the superiority of MAGI.





\section{Conclusion}

In this paper, we present MAGI, a cooperative MARL framework which introduces an explicit consensus mechanism for multi-agent coordination. 
Firstly, a distribution of future states is modeled with a novel CVAE-based self-supervised generative model. Then, a common goal with high potential value is sampled as multi-agent consensus to guide all agents' policies.
Experimental results on Multiagent Particle Environment and Google Research Football demonstrate the proposed consensus mechanism can effectively enhance the cooperation among agents and improve sample-efficiency.

\bibliography{ref}

\begin{thebibliography}{34}
\providecommand{\natexlab}[1]{#1}

\bibitem[{Ahilan and Dayan(2019)}]{ahilan2019feudal}
Ahilan, S.; and Dayan, P. 2019.
\newblock Feudal multi-agent hierarchies for cooperative reinforcement
  learning.
\newblock \emph{arXiv preprint arXiv:1901.08492}.

\bibitem[{Berner et~al.(2019)Berner, Brockman, Chan, Cheung, Debiak, Dennison,
  Farhi, Fischer, Hashme, Hesse et~al.}]{berner2019dota}
Berner, C.; Brockman, G.; Chan, B.; Cheung, V.; Debiak, P.; Dennison, C.;
  Farhi, D.; Fischer, Q.; Hashme, S.; Hesse, C.; et~al. 2019.
\newblock Dota 2 with large scale deep reinforcement learning.
\newblock \emph{arXiv preprint arXiv:1912.06680}.

\bibitem[{Cao et~al.(2012)Cao, Yu, Ren, and Chen}]{cao2012overview}
Cao, Y.; Yu, W.; Ren, W.; and Chen, G. 2012.
\newblock An overview of recent progress in the study of distributed
  multi-agent coordination.
\newblock \emph{IEEE Transactions on Industrial informatics}, 9(1): 427--438.

\bibitem[{David~Ha(2017)}]{ha2016hypernetworks}
David~Ha, Q. V.~L., Andrew M.~Dai. 2017.
\newblock HyperNetworks.
\newblock In \emph{5th International Conference on Learning Representations,
  {ICLR} 2017}.

\bibitem[{Gronauer and Diepold(2022)}]{gronauer2022multi}
Gronauer, S.; and Diepold, K. 2022.
\newblock Multi-agent deep reinforcement learning: a survey.
\newblock \emph{Artificial Intelligence Review}, 55(2): 895--943.

\bibitem[{H{\"u}ttenrauch, {\v{S}}o{\v{s}}i{\'c}, and
  Neumann(2017)}]{huttenrauch2017guided}
H{\"u}ttenrauch, M.; {\v{S}}o{\v{s}}i{\'c}, A.; and Neumann, G. 2017.
\newblock Guided deep reinforcement learning for swarm systems.
\newblock \emph{arXiv preprint arXiv:1709.06011}.

\bibitem[{Iqbal and Sha(2019)}]{iqbal2019actor}
Iqbal, S.; and Sha, F. 2019.
\newblock Actor-attention-critic for multi-agent reinforcement learning.
\newblock In \emph{International Conference on Machine Learning}, 2961--2970.

\bibitem[{Ke et~al.(2019)Ke, Singh, Touati, Goyal, Bengio, Parikh, and
  Batra}]{ke2019learning}
Ke, N.~R.; Singh, A.; Touati, A.; Goyal, A.; Bengio, Y.; Parikh, D.; and Batra,
  D. 2019.
\newblock Learning dynamics model in reinforcement learning by incorporating
  the long term future.
\newblock \emph{stat}, 1050: 16.

\bibitem[{Kingma and Welling(2014)}]{kingma2014auto}
Kingma, D.~P.; and Welling, M. 2014.
\newblock Auto-Encoding Variational Bayes.
\newblock \emph{stat}, 1050: 1.

\bibitem[{Konda and Tsitsiklis(2000)}]{konda2000actor}
Konda, V.~R.; and Tsitsiklis, J.~N. 2000.
\newblock Actor-critic algorithms.
\newblock In \emph{Advances in neural information processing systems},
  1008--1014.

\bibitem[{Krupnik, Mordatch, and Tamar(2020)}]{krupnik2020multi}
Krupnik, O.; Mordatch, I.; and Tamar, A. 2020.
\newblock Multi-agent reinforcement learning with multi-step generative models.
\newblock In \emph{Conference on Robot Learning}, 776--790.

\bibitem[{Lillicrap et~al.(2016)Lillicrap, Hunt, Pritzel, Heess, Erez, Tassa,
  Silver, and Wierstra}]{Lillicrap2016}
Lillicrap, T.~P.; Hunt, J.~J.; Pritzel, A.; Heess, N.; Erez, T.; Tassa, Y.;
  Silver, D.; and Wierstra, D. 2016.
\newblock {Continuous control with deep reinforcement learning}.
\newblock \emph{4th International Conference on Learning Representations, ICLR
  2016 - Conference Track Proceedings}.

\bibitem[{Littman(2001)}]{littman2001value}
Littman, M.~L. 2001.
\newblock Value-function reinforcement learning in Markov games.
\newblock \emph{Cognitive systems research}, 2(1): 55--66.

\bibitem[{Liu et~al.(2021)Liu, Jain, Yeh, and Schwing}]{liu2021cooperative}
Liu, I.-J.; Jain, U.; Yeh, R.~A.; and Schwing, A. 2021.
\newblock Cooperative exploration for multi-agent deep reinforcement learning.
\newblock In \emph{International Conference on Machine Learning}, 6826--6836.
  PMLR.

\bibitem[{Lowe et~al.(2017)Lowe, Wu, Tamar, Harb, Abbeel, and
  Mordatch}]{lowe2017multi}
Lowe, R.; Wu, Y.~I.; Tamar, A.; Harb, J.; Abbeel, O.~P.; and Mordatch, I. 2017.
\newblock Multi-agent actor-critic for mixed cooperative-competitive
  environments.
\newblock In \emph{Advances in neural information processing systems},
  6379--6390.

\bibitem[{Matignon, Laurent, and Le~Fort-Piat(2012)}]{matignon2012independent}
Matignon, L.; Laurent, G.~J.; and Le~Fort-Piat, N. 2012.
\newblock Independent reinforcement learners in cooperative markov games: a
  survey regarding coordination problems.
\newblock \emph{The Knowledge Engineering Review}, 27(1): 1--31.

\bibitem[{Mishra, Abbeel, and Mordatch(2017)}]{mishra2017prediction}
Mishra, N.; Abbeel, P.; and Mordatch, I. 2017.
\newblock Prediction and control with temporal segment models.
\newblock In \emph{International Conference on Machine Learning}, 2459--2468.
  PMLR.

\bibitem[{Nagabandi et~al.(2018)Nagabandi, Kahn, Fearing, and
  Levine}]{nagabandi2018neural}
Nagabandi, A.; Kahn, G.; Fearing, R.~S.; and Levine, S. 2018.
\newblock Neural network dynamics for model-based deep reinforcement learning
  with model-free fine-tuning.
\newblock In \emph{IEEE International Conference on Robotics and Automation},
  7559--7566. IEEE.

\bibitem[{Omidshafiei et~al.(2017)Omidshafiei, Pazis, Amato, How, and
  Vian}]{omidshafiei2017deep}
Omidshafiei, S.; Pazis, J.; Amato, C.; How, J.~P.; and Vian, J. 2017.
\newblock Deep decentralized multi-task multi-agent reinforcement learning
  under partial observability.
\newblock In \emph{International Conference on Machine Learning}, 2681--2690.
  PMLR.

\bibitem[{Peng et~al.(2021)Peng, Rashid, Schroeder~de Witt, Kamienny, Torr,
  B{\"o}hmer, and Whiteson}]{peng2021facmac}
Peng, B.; Rashid, T.; Schroeder~de Witt, C.; Kamienny, P.-A.; Torr, P.;
  B{\"o}hmer, W.; and Whiteson, S. 2021.
\newblock Facmac: Factored multi-agent centralised policy gradients.
\newblock \emph{Advances in Neural Information Processing Systems}, 34:
  12208--12221.

\bibitem[{Racani{\`e}re et~al.(2017)Racani{\`e}re, Weber, Reichert, Buesing,
  Guez, Rezende, Badia, Vinyals, Heess, Li et~al.}]{racaniere2017imagination}
Racani{\`e}re, S.; Weber, T.; Reichert, D.~P.; Buesing, L.; Guez, A.; Rezende,
  D.; Badia, A.~P.; Vinyals, O.; Heess, N.; Li, Y.; et~al. 2017.
\newblock Imagination-augmented agents for deep reinforcement learning.
\newblock In \emph{Proceedings of the 31st International Conference on Neural
  Information Processing Systems}, 5694--5705.

\bibitem[{Rashid et~al.(2018)Rashid, Samvelyan, Schroeder, Farquhar, Foerster,
  and Whiteson}]{rashid2018qmix}
Rashid, T.; Samvelyan, M.; Schroeder, C.; Farquhar, G.; Foerster, J.; and
  Whiteson, S. 2018.
\newblock QMIX: Monotonic Value Function Factorisation for Deep Multi-Agent
  Reinforcement Learning.
\newblock In \emph{International Conference on Machine Learning}, 4295--4304.

\bibitem[{Ren, Beard, and Atkins(2005)}]{ren2005survey}
Ren, W.; Beard, R.~W.; and Atkins, E.~M. 2005.
\newblock A survey of consensus problems in multi-agent coordination.
\newblock In \emph{Proceedings of the 2005, American Control Conference,
  2005.}, 1859--1864. IEEE.

\bibitem[{Schulman et~al.(2017)Schulman, Wolski, Dhariwal, Radford, and
  Klimov}]{schulman2017proximal}
Schulman, J.; Wolski, F.; Dhariwal, P.; Radford, A.; and Klimov, O. 2017.
\newblock Proximal policy optimization algorithms.
\newblock \emph{arXiv preprint arXiv:1707.06347}.

\bibitem[{Sohn, Lee, and Yan(2015)}]{sohn2015learning}
Sohn, K.; Lee, H.; and Yan, X. 2015.
\newblock Learning structured output representation using deep conditional
  generative models.
\newblock \emph{Advances in neural information processing systems}, 28:
  3483--3491.

\bibitem[{Son et~al.(2019)Son, Kim, Kang, Hostallero, and Yi}]{son2019qtran}
Son, K.; Kim, D.; Kang, W.~J.; Hostallero, D.~E.; and Yi, Y. 2019.
\newblock QTRAN: Learning to Factorize with Transformation for Cooperative
  Multi-Agent Reinforcement Learning.
\newblock In \emph{International Conference on Machine Learning}, 5887--5896.

\bibitem[{Sunehag et~al.(2018)Sunehag, Lever, Gruslys, Czarnecki, Zambaldi,
  Jaderberg, Lanctot, Sonnerat, Leibo, Tuyls et~al.}]{sunehag2018value}
Sunehag, P.; Lever, G.; Gruslys, A.; Czarnecki, W.~M.; Zambaldi, V.~F.;
  Jaderberg, M.; Lanctot, M.; Sonnerat, N.; Leibo, J.~Z.; Tuyls, K.; et~al.
  2018.
\newblock Value-Decomposition Networks For Cooperative Multi-Agent Learning
  Based On Team Reward.
\newblock In \emph{International Conference On Autonomous Agents and
  Multi-Agent Systems}, 2085--2087.

\bibitem[{Sutton(1991)}]{sutton1991dyna}
Sutton, R.~S. 1991.
\newblock Dyna, an integrated architecture for learning, planning, and
  reacting.
\newblock \emph{ACM Sigart Bulletin}, 2(4): 160--163.

\bibitem[{Sutton and Barto(2018)}]{sutton2018reinforcement}
Sutton, R.~S.; and Barto, A.~G. 2018.
\newblock \emph{Reinforcement learning: An introduction}.
\newblock MIT press.

\bibitem[{Talvitie(2014)}]{talvitie2014model}
Talvitie, E. 2014.
\newblock Model regularization for stable sample rollouts.
\newblock In \emph{Proceedings of the Thirtieth Conference on Uncertainty in
  Artificial Intelligence}, 780--789.

\bibitem[{Tampuu et~al.(2017)Tampuu, Matiisen, Kodelja, Kuzovkin, Korjus, Aru,
  Aru, and Vicente}]{tampuu2017multiagent}
Tampuu, A.; Matiisen, T.; Kodelja, D.; Kuzovkin, I.; Korjus, K.; Aru, J.; Aru,
  J.; and Vicente, R. 2017.
\newblock Multiagent cooperation and competition with deep reinforcement
  learning.
\newblock \emph{PloS one}, 12(4): e0172395.

\bibitem[{Vezhnevets et~al.(2017)Vezhnevets, Osindero, Schaul, Heess,
  Jaderberg, Silver, and Kavukcuoglu}]{vezhnevets2017feudal}
Vezhnevets, A.~S.; Osindero, S.; Schaul, T.; Heess, N.; Jaderberg, M.; Silver,
  D.; and Kavukcuoglu, K. 2017.
\newblock Feudal networks for hierarchical reinforcement learning.
\newblock In \emph{International Conference on Machine Learning}, 3540--3549.

\bibitem[{Ye et~al.(2020)Ye, Chen, Zhang, Chen, Yuan, Liu, Chen, Liu, Qiu, Yu
  et~al.}]{ye2020towards}
Ye, D.; Chen, G.; Zhang, W.; Chen, S.; Yuan, B.; Liu, B.; Chen, J.; Liu, Z.;
  Qiu, F.; Yu, H.; et~al. 2020.
\newblock Towards Playing Full MOBA Games with Deep Reinforcement Learning.
\newblock \emph{arXiv e-prints}, arXiv--2011.

\bibitem[{Ye, Zhang, and Yang(2015)}]{ye2015multi}
Ye, D.; Zhang, M.; and Yang, Y. 2015.
\newblock A multi-agent framework for packet routing in wireless sensor
  networks.
\newblock \emph{sensors}, 15(5): 10026--10047.

\end{thebibliography}

\newpage
\section*{Appendix}



\section*{A. Generalized Intrinsic Reward} \label{sec:latent_space_reward}
In Section 5, we implement intrinsic reward in Equation \ref{eq:intrinsic_reward} by calculating the Euclidean distance between agent and goal positions.
In fact, the intrinsic reward can be calculated without any domain knowledge.
Here we provide a generalized implementation of intrinsic reward, which makes \MN can be used in any complex environment.

\begin{figure}[ht]
	\centering
	\includegraphics[width=85mm]{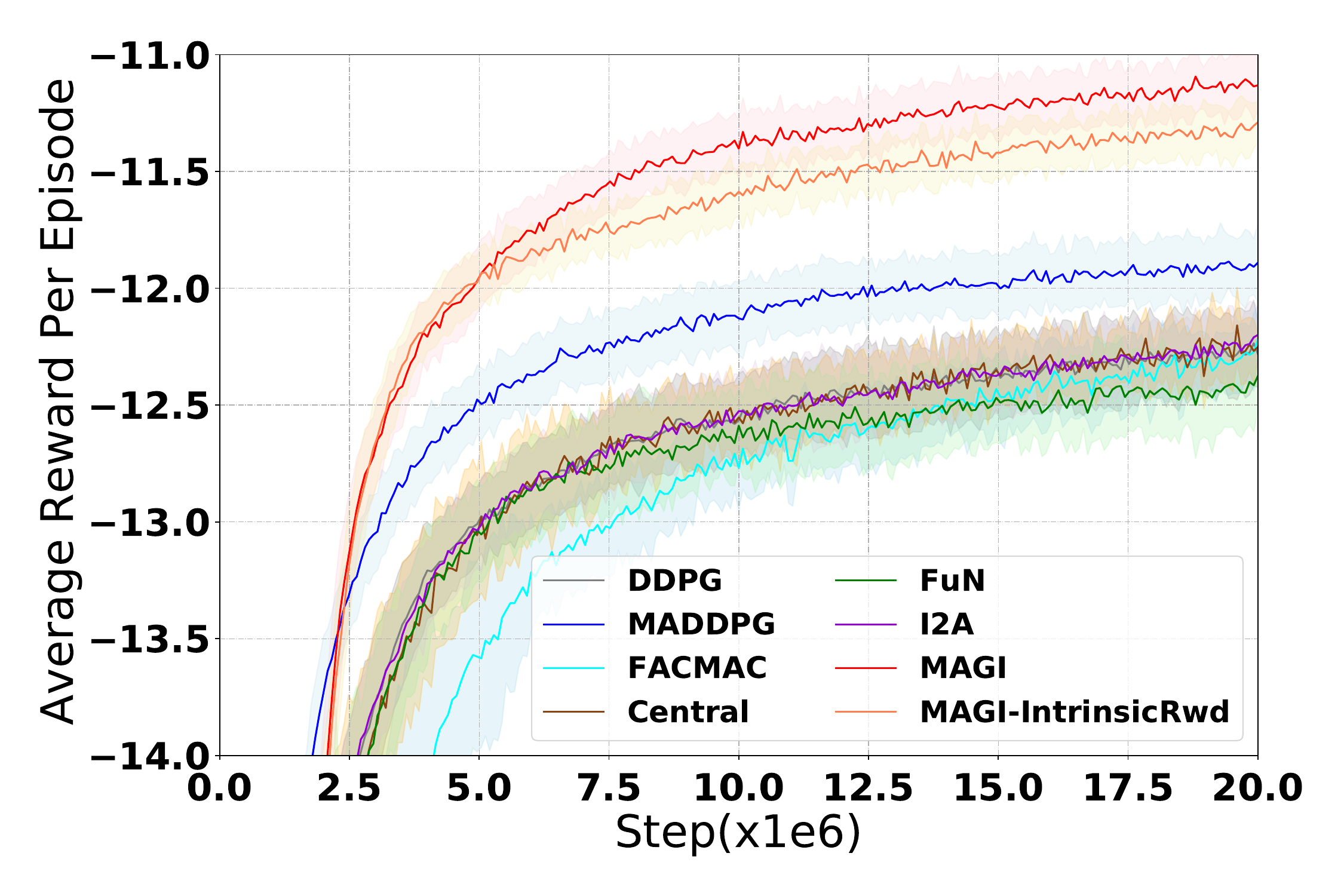}
	\caption{Results of MAGI with generalized intrinsic reward.}
	\label{fig:latent_space_reward}
\end{figure}

In this generalized version, the agent state and the goal state are mapped to CVAE latent space to calculate the difference.
Specifically, in Equation \ref{eq:intrinsic_reward},  $f(\cdot)$ and $g(\cdot)$ are implemented with the posterior distribution in CVAE $h_{\theta^{\text{enc}}}(\cdot)=q_{\theta^{\text{enc}}}(z|\cdot,s_t)$, and their distance is calculated with KL divergence.
Thus the intrinsic reward becomes:
\begin{equation*}
	\begin{aligned}
		r^{i\text{-in}}_{t} = {\mathbb{KL}}[h_{\theta^{\text{enc}}}(s_t^g) \| h_{\theta^{\text{enc}}}(s_{t})] - {\mathbb{KL}}[h_{\theta^{\text{enc}}}(s_t^g) \| h_{\theta^{\text{enc}}}(s_{t+1})]
	\end{aligned}\label{eq:latent_space_reward}
\end{equation*}
This distance metric contains high-dimensional information about the difference between $s_t^g$ and $s_t$ and can be used in any environment without domain knowledge.

We implemented MAGI with the above variant $r^{i\text{-in}}$ in the MPEs Navigation scenario.
As shown in Figure \ref{fig:latent_space_reward}, MAGI can achieve competitive performance compared to the version presented in Section 5.

\section*{B. Reward Designs in Multi-agent Particle-Environments}
All agents share the same reward.

\textit{Navigation} Agents are rewarded using sum of minimum distance to any agents for each landmark at each time step. Agents get -1 when they collide with each other. Total time steps for one episode game is 25.

\textit{Treasure Collection} Agents get +1  when they collect any treasure and get -1 when they collide with each other.  Total time steps for one episode game is 25.

\textit{Predator-Prey} Agents get +10  they collide with the prey. Total time steps for one episode game is 100.

\textit{Keep Away} Agents get +1  when they collide with the theft agent and get -1 when they collide with each other.  Total time steps for one episode game is 100.

\end{document}